\pdfoutput=1

\documentclass[11pt]{article}

\usepackage{graphicx}
\usepackage{subcaption}
\usepackage{booktabs}
\usepackage{multirow}
\usepackage{lscape}
\usepackage{natbib} %
\usepackage{todonotes}

\newcommand{\titletext}{A Little Human Data Goes A Long Way}

\usepackage{amsmath,amsfonts,bm}

\def\eqref#1{equation~\ref{#1}}

\def\1{\bm{1}}

\DeclareMathAlphabet{\mathsfit}{\encodingdefault}{\sfdefault}{m}{sl}
\SetMathAlphabet{\mathsfit}{bold}{\encodingdefault}{\sfdefault}{bx}{n}

\usepackage[final]{acl}

\usepackage{times}
\usepackage{latexsym}

\usepackage{cleveref} 

\usepackage[T1]{fontenc}

\usepackage[utf8]{inputenc}

\usepackage{microtype}

\usepackage{inconsolata}

\usepackage{graphicx}

\title{\titletext}

\author{Dhananjay Ashok\textsuperscript{$\dagger$},  \quad
  Jonathan May\textsuperscript{$\dagger$} \\
  \textsuperscript{$\dagger$}Information Sciences Institute, University of Southern California\\
  \texttt{\{ashokd, jonmay\}@isi.edu} \\}

\begin{document}
\maketitle
\begin{abstract}
Faced with an expensive human annotation process, creators of NLP systems increasingly turn to synthetic data generation. While this method shows promise, the extent to which synthetic data can replace human annotation is poorly understood. We investigate the use of synthetic data in Fact Verification (FV) and Evidence-based Question Answering (QA) by incrementally replacing human-generated data with synthetic points on eight diverse datasets. Strikingly, replacing up to 90\% of the training data only marginally decreases performance, but replacing the final 10\% leads to severe declines. We find that models trained on purely synthetic data can be improved by including as few as \textbf{125} human generated data points. We show that matching the performance gain of a little human data requires an order of magnitude more synthetic data, and then estimate price ratios at which human annotation would be a more cost-effective solution. Our results suggest that even when human annotation at scale is infeasible, there is great value to having a small proportion of the dataset being human-generated.

\begin{figure}[t]   \includegraphics[width=0.45\textwidth]{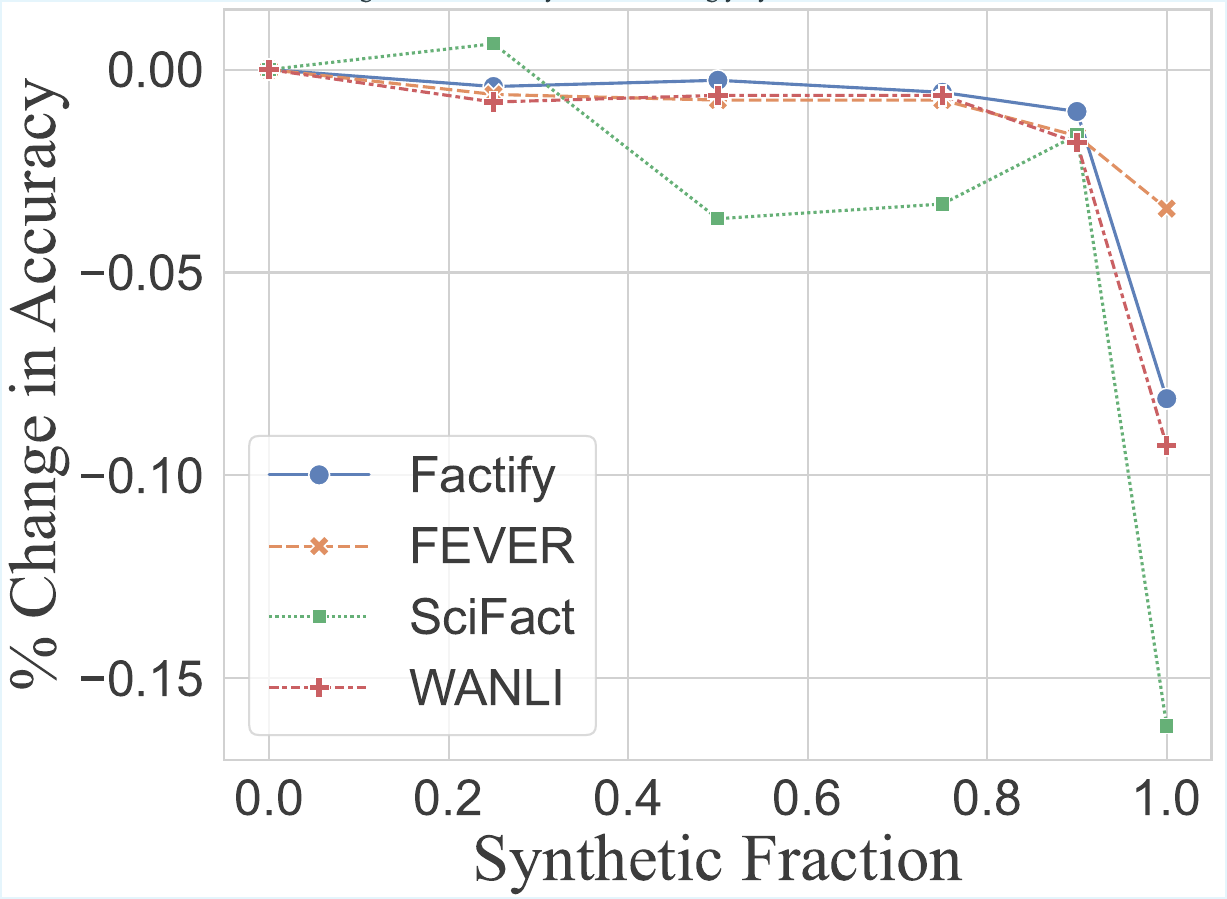}
\includegraphics[width=0.45\textwidth]{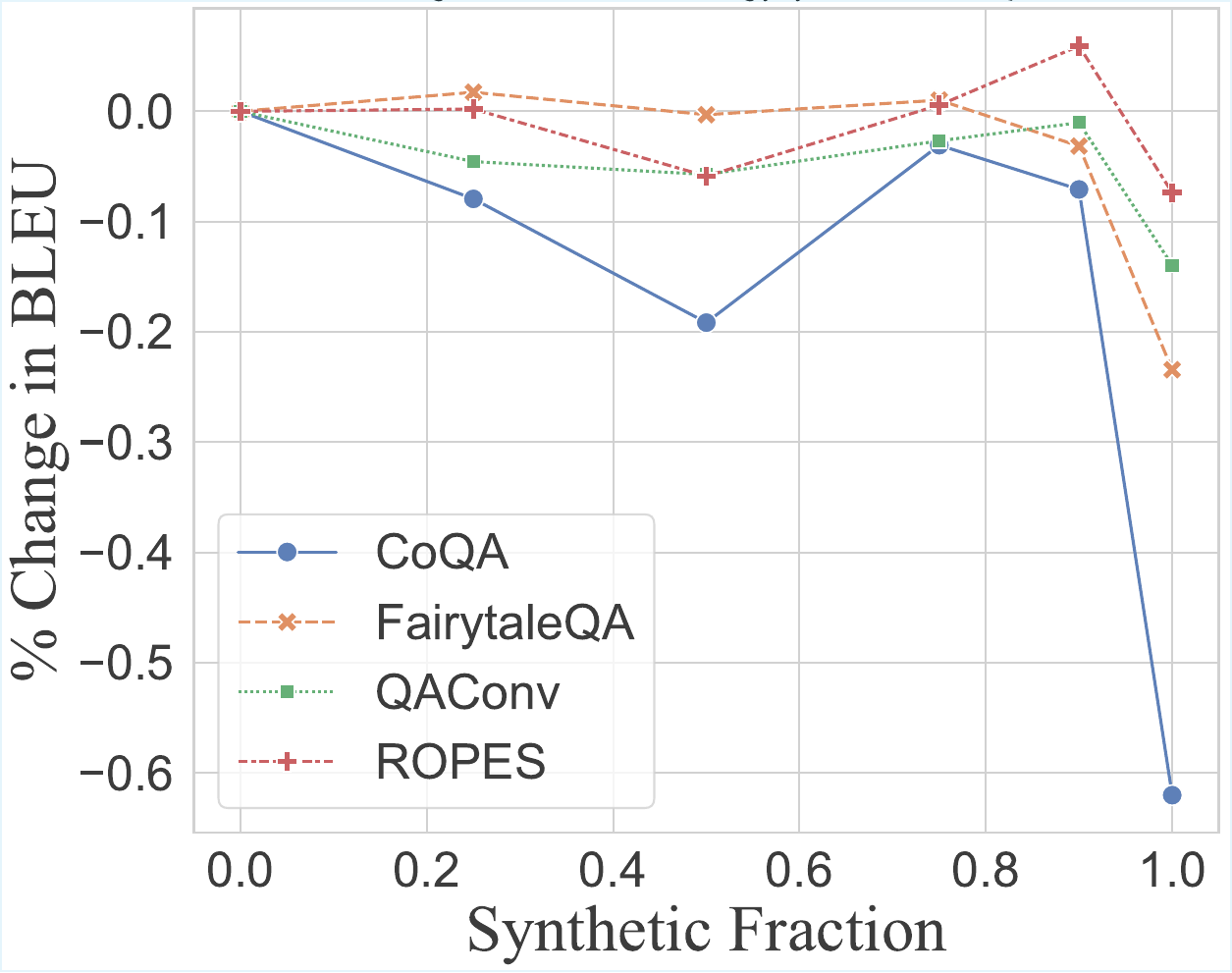}
        \caption{Change in model performance as the proportion of synthetic points in the training data is increased. Across datasets, the performance decrease when moving the synthetic proportion from 0 to 0.90 is often less than that of moving from 0.9 to purely synthetic data.}
        \label{fig:full}
\end{figure}

\end{abstract}

\section{Introduction}
\label{sec:introduction}
From BERT \citep{devlin-etal-2019-bert} to GPT-4 \citep{achiam2023gpt}, the explosive growth of language models (LMs) has been underpinned by exponential increases in the size of available training data. However, the more complex and specialized the task, the more expensive and challenging it is to collect human-generated data at scale~\citep{wang-etal-2021-want-reduce}. Combined with growing concerns that LMs may soon exhaust the stock of publicly available training data~\citep{Villalobos2022WillWR}, many turn to synthetic data generation, hoping to eliminate their reliance on human annotation.

Synthetic data generation has long been used to increase the amount of training data available~\cite{simard2002transformation,krizhevsky2012imagenet}. Early NLP approaches use rule based methods~\citep{de2005improving, chen2012prefer}, paraphrasing~\citep{Wang2015ThatsSA, Kobayashi2018ContextualAD}, noising~\citep{Xie2017DataNA, Wang2018SwitchOutAE}, and backtranslation~\citep{Sennrich2015ImprovingNM, Yu2018QANetCL}, but are limited in their capability. 

Modern LMs demonstrate the capability to solve myriad NLP tasks with minimal task specific data~\citep{brown2020language, wei2022emergent}, making them more powerful synthetic data generators. Leveraging this, synthetic data approaches have seen increased use in tasks~\citep{Tan2024LargeLM} such as QA~\citep{Wu2021QAConvQA}, natural language inference (NLI)~\citep{Meng2022GeneratingTD}, text classification~\citep{Ye2022ZeroGenEZ}, instruction tuning~\citep{Li2023SelfAlignmentWI},  evaluation~\citep{Dubois2023AlpacaFarmAS},  and more~\citep{Tang2023DoesSD}.

\begin{figure}[t]   \includegraphics[width=0.48\textwidth]{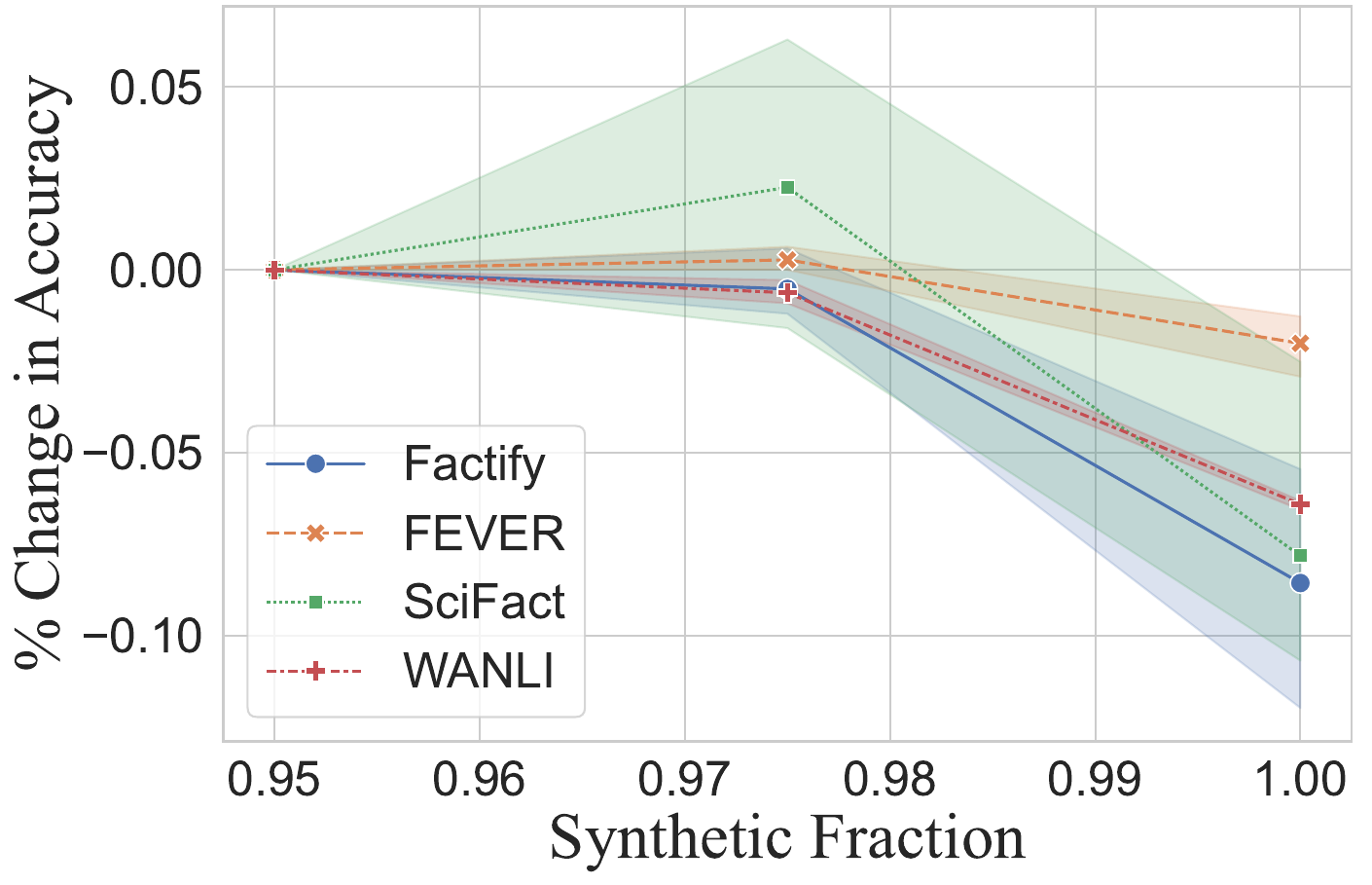}
\includegraphics[width=0.48\textwidth]{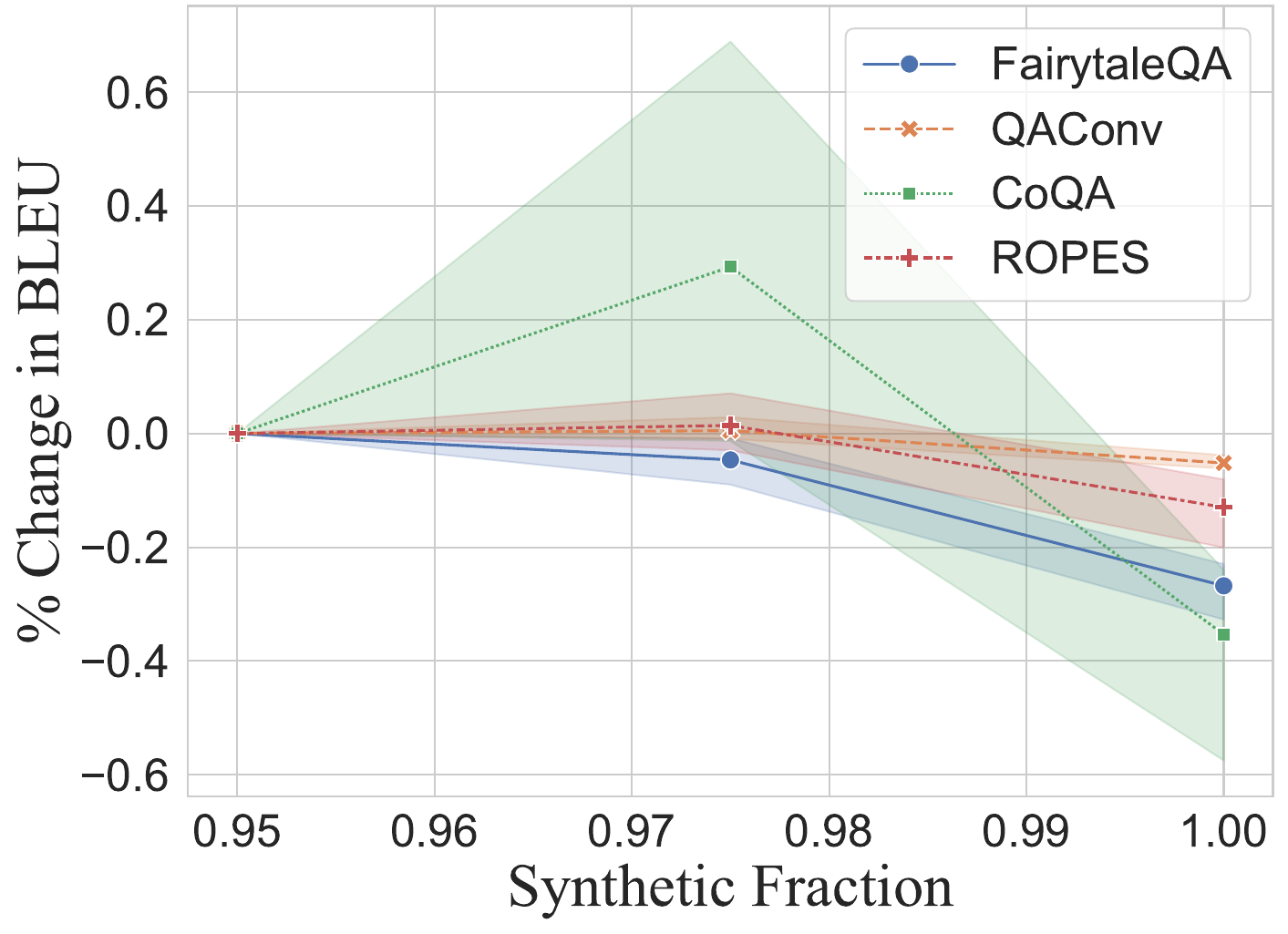}
        \caption{Model performance as the synthetic proportion of the training data varies from 0.95 to 1. Having \textbf{just 2.5\%} of the training dataset being human-generated boosts performance.}
        \label{fig:zoom5joined}
\end{figure}

The adoption has been particularly enthusiastic for tasks that require the model to `understand' knowledge contained in an `evidence text' e.g., FV~\citep{Tang2024MiniCheckEF}, factual error correction~\citep{ashok2023scifix}, NLI~\citep{Hosseini2024ASD} and evidence-based QA~\citep{Schimanski2024TowardsFA}. Such tasks are of vital importance in fake news detection~\citep{Sharma2023FakeND}, retrieval-augmented generation~\citep{Gao2023RetrievalAugmentedGF} and dialogue systems~\citep{Weston2015TowardsAQ}.  Recent datasets~\citep{Wu2021QAConvQA} and methods~\citep{Ye2022ZeroGenEZ} exploit plentiful evidence texts (scientific journals, news articles, books, etc.), using synthetic generation to avoid being bottlenecked by the expensive annotation procedure~\citep{Liu2022WANLIWA}.

Varying results across ML tasks suggest that whether completely replacing humans with synthetic data shows promise~\citep{Fan2023ScalingLO, Hammoud2024SynthCLIPAW} or leads to failures~\citep{Bisbee2024SyntheticRF, Guo2023TheCD} is task dependent~\citep{Li2023SyntheticDG}. In this work, we focus on \textbf{FV} and \textbf{Evidence-based QA}, performing the first investigation into the trade-offs presented by the use of synthetic data generation in these fundamental tasks.

We study eight diverse FV and QA datasets, using their `evidence texts' to generate synthetic datasets. By holding the number of data points constant but increasing the percentage of the training data that is synthetic, we can compare the utility of synthetic data to the original human generated data points. Across multiple models, prompt models, and prompting strategies, we find (Figure~\ref{fig:full}) that while increasing the proportion of synthetic data typically causes only minor degradations in model performance, a significant decline occurs at the extremes; i.e., when the percentage of synthetic data exceeds 90\%. Focusing on the extremes, we show that purely synthetically trained FV and QA systems can be meaningfully improved by including as few as \textbf{125} human-generated datapoints.

Our observations have actionable implications for researchers hoping to use synthetic data for FV and QA. The results (Figure~\ref{fig:zoom5joined}, Figure~\ref{fig:money_main}) suggest that even when human annotation at scale is infeasible, there is great value to having a small proportion of the dataset being generated by humans. 

To help guide this choice, we quantify the performance-cost tradeoff between human and synthetic data. We find (Figure~\ref{fig:money_main}) that matching the performance gain of just a little additional human data (only 200 data points) requires an order of magnitude more synthetic data points, empirically showing the per-data point price ratio at which human annotation is the more cost-effective solution. 
Finally, we conduct an analysis on the differing properties of synthetic vs. human data. Among other findings, we see that synthetic generations are longer and more extractive from the evidence texts than their human-produced counterparts.

\section{Synthetic Data Generation from Evidence Texts}
\label{sec:synthetic}
We study a synthetic generation pipeline representative of the methods used in the FV~\citep{Ni2024AFaCTAAT, He2022IsSD} and QA~\citep{Schimanski2024TowardsFA, Wan2024SciQAGAF} literature.
Using Few-Shot In-Context Learning~\citep{brown2020language}, we generate synthetic (claim, label) pairs from an input evidence text. The prompt model is given examples of (evidence text, claim, label) from the human training data, and is then queried with the evidence text we seek to generate data for. QA synthetic data is generated analogously, see details in  Appendix~\ref{sec:appendix_synthetic}.

In total, we use four FV/NLI datasets: FEVER~\citep{thorne2018fever}, SciFact~\citep{wadden2020fact}, WANLI~\citep{Liu2022WANLIWA} and FACTIFY1.0~\citep{Mishra2022FACTIFYAM}, as well as four QA datasets: ROPES~\citep{Lin2019ReasoningOP}, CoQA~\citep{reddy2019coqa}, QAConv~\citep{Wu2021QAConvQA} and FairyTaleQA~\citep{xu2022fantastic}. Together, the datasets span a variety of domains (science, news, social media, reasoning, conversation, fiction). We confirm that the generations are of high quality by verifying that the diversity of the synthetic data is comparable to the human generated samples. For more details, including a discussion on data leakage, see  Appendix~\ref{sec:appendix_datasets}. 

FV performance is measured by test accuracy, while QA is measured using BLEU~\citep{papineni-etal-2002-bleu}; we show robustness to choice of metric in Appendix~\ref{sec:supplement}. Evaluation is always conducted on the (human-generated) test split of each dataset. 

We use GPT-3.5-Turbo~\citep{brown2020language} for prompting and LoRA~\citep{hu2021lora} on Llama3-8B~\citep{dubey2024llama} for fine-tuning. Implementation details are provided in Appendix~\ref{sec:appendix_implementation}, and our code is publicly available~\footnote{\url{https://github.com/DhananjayAshok/LittleHumanData/}}

\begin{figure}[t]   \includegraphics[width=0.48\textwidth]{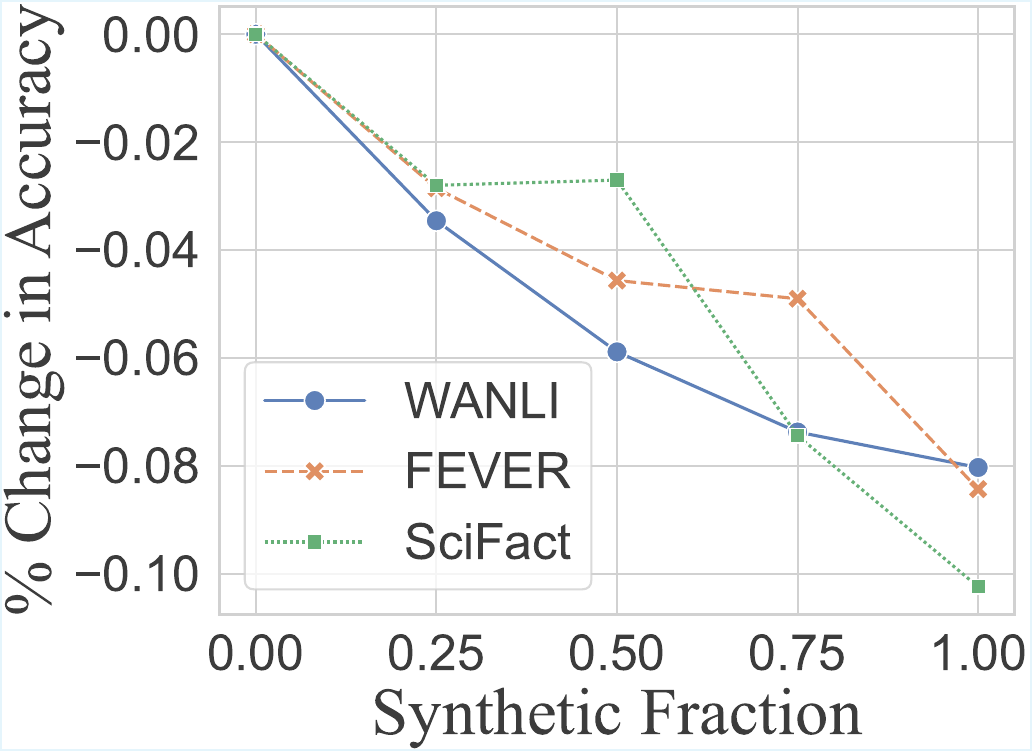}
        \caption{Change in accuracy when the test set (shown in key) is not seen during training, and the training set is a mixture of other FV datasets. Increasing the synthetic proportion of the dataset leads to performance declines even in the OOD setting, showing that human data offers genuine performance increases.}
        \label{fig:ood}
\end{figure}

\section{Can Synthetic Data Replace Humans?}
\label{sec:frac}
We investigate the potential of synthetic data to replace human annotation by holding the number of training data points fixed, incrementally increasing the proportion of the data that is synthetic, and fine-tuning a model on each training set.

\begin{figure}[t]
    \includegraphics[width=0.48\textwidth]{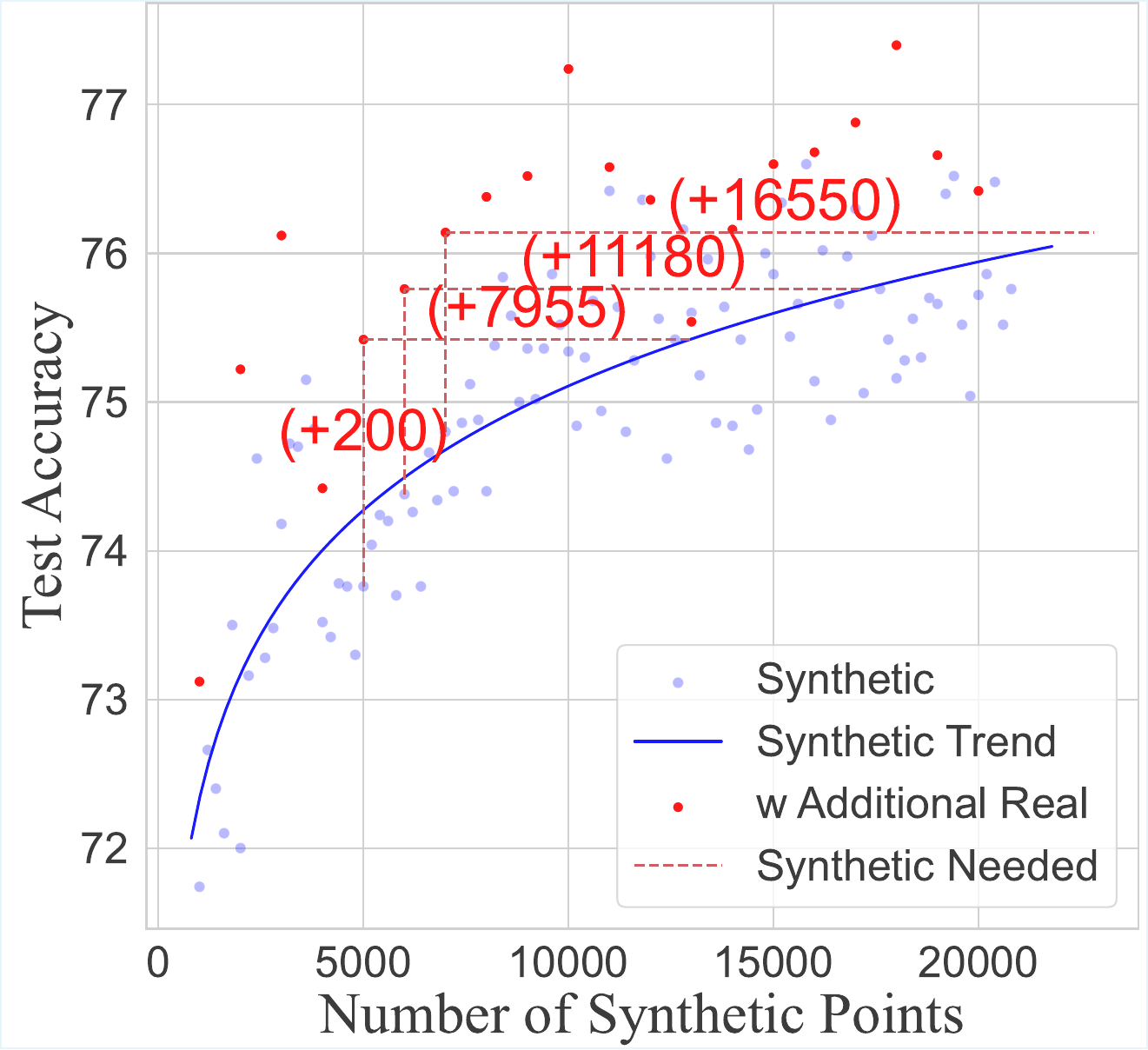}
        \caption{On the WANLI dataset, adding 200 real data points is as effective as adding an order of magnitude more synthetic data points.}
        \label{fig:money_main}
\end{figure}

\noindent\textbf{Results:}
Across all datasets, using purely synthetic data leads to worse performance than the same amount of human data (Figure~\ref{fig:full}). 
We consider the possibility that this result could be caused by a spurious correlation between the human training and testing splits (e.g., annotation artifacts that are correlated with the label but not fundamental to the task).  We conduct an out-of-distribution experiment, using different datasets for training and testing (e.g., training on FEVER + SciFACT and testing on WANLI). Increasing the synthetic proportion leads to performance declines even in the OOD setting (Figure~\ref{fig:ood}), showing that human data offers genuine performance increases, and the results cannot be explained by a spurious correlation between the human test and human training samples (further discussion in Appendix~\ref{sec:supplement}).

The performance decline is not uniform as we increase the synthetic proportion. On almost all datasets, there is only a minor degradation up until 90\% replacement, after which the performance drops considerably. We zoom in on the 90\%-100\% interval, fixing the amount of training data at $n=5000$ (500 for SciFact) and training on datasets with 95\%, 97.5\% and 100\% synthetic data (Figure~\ref{fig:zoom5joined}).
Surprisingly, the results show that there is a significant difference between the performance of models on 97.5\% and 100\% synthetic data; the addition of just \textbf{125} (2.5\% of 5000) human generated datapoints reliably improves the performance of synthetically trained FV and QA models. These trends hold robustly over different languages (Arabic, Georgian, Indonesian), choice of fine-tuning model (Mistral, MPT), prompt model (GPT4 and Claude-3.5-Sonnet), prompting strategy (Chain-of-Thought), model size and dataset size (Appendix~\ref{sec:supplement}).

\section{When Should We Use Human Data?}
\label{sec:exp3}
\begin{table*}[th]
\centering
\begin{tabular}{lrrrr}
\hline
\textbf{Dataset} & \textbf{Mean} & \textbf{Median} & \textbf{25th Percentile} & \textbf{75th Percentile} \\ \hline
WANLI            & 17,671       & 16,905        & 9,711                  & 22,931                 \\
ROPES            & 17,333      & 6,006         & 3,623                  & 21,944                 \\
FairyTaleQA      & 281,951     & 36,901        & 15,129                 & 813,135                \\
FEVER            & 1,155       & 237          & -1,400                 & 7,073                  \\ \hline
\end{tabular}
\caption{Additional synthetic data points needed to match the performance gain of 200 human data points. High values for FairyTaleQA suggest that human-generated data may unlock performance that purely synthetic data cannot achieve. Negative values for FEVER are due to a saturation of the performance gains, however, human data points reach the saturation point much faster (Appendix~\ref{sec:supplement})}.
\label{table:money}
\end{table*}
Having observed the disproportionate value added by human data, we ask what the relative cost between human and synthetic data generation must be for us to prefer one over the other. We fine-tune models on purely synthetic datasets of varying sizes, and establish the synthetic baseline by fitting a curve of the form $y = a_0 + a_1\log(x)$ where $x$ is the size of the synthetic dataset and $y$ is the performance. We then take the synthetic training sets with \{1000, 2000 \ldots\} points and observe the performance ($y^*$) when we add 200 human data points. $\exp({\frac{y^*-a_0}{a1}})$ is then the size of the purely synthetic dataset that achieves equivalent performance.  

\begin{table}[t]
\centering
\begin{tabular}{@{}lrr@{}}
\toprule
Dataset     & Synthetic  & Human \\ \midrule
FEVER       & \textbf{5.50}                        & 20.27                  \\
WANLI       & \textbf{0.23}                       & 1.22                   \\
SCIFACT     & \textbf{0.00}                          & 9.93                   \\
FACTIFY     & \textbf{0.27}                       & 14.93                  \\ \midrule
NarrativeQA & 3.85                       & \textbf{1.42}                   \\
CoQA        & \textbf{0.54}                       & 5.49                   \\
FairyTaleQA & 2.26                       & \textbf{0.18}                   \\
ROPES       & 2.40                        & \textbf{1.35}                   \\ \bottomrule
\end{tabular}
\caption{Percentage of duplicated claims/questions for synthetic v.s. human data. Rates are comparable across datasets, but for fact verification datasets, synthetic datasets have fewer duplicates.}
\label{table:duplicates}
\end{table}

\noindent\textbf{Results:}
Across all datasets, adding \textbf{200} human data points is usually comparable to adding at least an order of magnitude (often multiple orders of magnitude) more synthetic data points. On WANLI (Figure~\ref{fig:money_main}), more than $17,000$ additional synthetic points are needed to achieve the performance gains of $200$ human points. If the price of a synthetic point for WANLI exceeds $73$ times the price of a human generated point, then an incremental amount of human annotation would be a more cost-effective way to achieve the same increase in accuracy. In the extreme case, the equation learned on FairyTaleQA suggests that it takes $2e5$ additional synthetic points to match the performance gain of 200 additional human data points. Rather than interpret these numbers literally, we take them to suggest that human data could have unique value in some settings, enabling performance levels that are impossible with purely synthetic datasets. See Appendix~\ref{sec:supplement} for more results and details.

\section{Discussion}
\label{sec:discussion}
\begin{figure}[t]
     \centering
\includegraphics[width=0.5\textwidth]{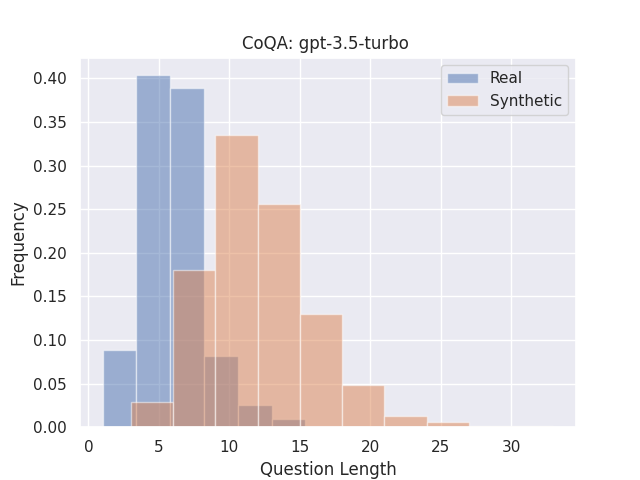}
        \caption{Synthetic questions are longer than human generated ones, a trend also seen in answers.}
        \label{fig:size_main}
\end{figure}
The synthetic generations are  as diverse as human data (Table~\ref{table:duplicates}), with comparable duplicate rates on QA datasets, and markedly fewer duplicates on most FV datasets. This is evidence that our synthetic generations are of good quality, however, even on the datasets where the synthetic data is significantly more diverse, the synthetic data does not perform as well. This suggests that diversity is an insufficient measure of quality when evaluating how good the generated data is.
Our analysis shows that synthetic data generation produces claims of comparable length to the human datasets, however synthetic questions and answers tend to be longer than human-generated counterparts for all QA datasets (Figure~\ref{fig:size_main}). We find that synthetic generations have a higher n-gram overlap with the evidence sentences. This suggests that synthetic data generation produces data points that are more directly taken from the evidence texts, while humans are more likely to employ rephrasing or different vocabulary than the evidence texts. Surprisingly, we find that synthetic data generation chooses more varied parts of the input text as sources for the question and answer content, with human annotation overwhelmingly more likely to create questions whose answers lie at the start of the evidence texts. We include a detailed discussion in Appendix~\ref{sec:appendix_dicussion}.

\section{Related Work}
\label{sec:related-work}
The replacement of human annotation with synthetic data is extensively studied in the pretraining stage of LMs, where results consistently show~\citep{Shumailov2023TheCO, Seddik2024HowBI, Guo2023TheCD, Briesch2023LargeLM} catastrophic forgetting, mode collapse, and performance deterioration.

In our setting, relying only on synthetic data still achieves reasonable performance across all tasks. This suggests that the usage of exclusively synthetic data poses fewer risks when generations are grounded in diverse, natural `evidence texts.' 

Interestingly, conclusions which confirm our findings are found more in the image and multimodal domains, where recent work~\citep{Singh2024IsSD, He2022IsSD, Fan2023ScalingLO} finds that synthetic data holds promise, but must be used in conjunction with human data to mitigate its harms. 

There is limited work on understanding whether synthetic data can replace human annotation in a task-specific setting for the language domain.~\citet{Li2023SyntheticDG} categorize text classification tasks by subjectivity, showing that synthetic data is less useful when tasks are more subjective. This draws them to focus on different tasks (sentiment classification, relation extraction and spam detection), and they do not study using a mixture of real and synthetic data.  
~\citet{Bisbee2024SyntheticRF} demonstrate that replacing political survey respondents with LMs produces unreliable results, while ~\citet{Ahmed2024CanLR} find that there are specific software engineering subtasks where synthetic data approaches human performance. ~\citet{Chen2024UnveilingTF} show that instruction-following capabilities are diminished when using synthetic data and present a machine unlearning approach to mitigate this. The diversity of results when evaluating the impact of using purely synthetic data confirms that the feasibility of replacing human annotation with synthetic data is highly task dependant. This work deepens our understanding of the problem by being the first to study whether synthetic data can replace human annotation on the fundamental tasks of fact verification and evidence-based question answering.

\section{Conclusion}
\label{sec:conclusion}
Showing impressive performance when human data is scarce, synthetic data generation seems poised to remain a key method in FV and QA. Our work sheds light on how the best way to use this method is in conjunction with human data. We show that a little human data goes a long way, with just \textbf{125} points being enough to see reliable gains on all datasets studied. With practical considerations in mind, we show that the alternative to small amounts of additional human data can be an order of magnitude more of synthetic data, suggesting that at times human annotation can be cost-effective relative to synthetic generation. We hope these results better inform design decisions on datasets and methods for fact verification and question answering.

\section{Limitations}
\label{sec:limitations}
While we include results on multilingual Fact Verification datasets, the primary focus of our work is limited to the English language. Additionally, our results on  multilingual datasets suggest that while similar claims can be made regarding the impact of replacing human annotation with synthetic data across different languages, the amount of human data needed to observe a meaningful performance increase may vary across languages. We also have a limited ability to control for dataset leakage, with only one dataset from each of the tasks that is surely not leaked to GPT-3.5 (and, even these two datasets may have been seen by GPT-4). This can potentially bias the results in favor of synthetic data. Due to the scarcity of suitable available datasets (i.e., ones that have not been exposed to the prompt models) we are prevented from studying the problem more rigorously. Another limitation is that while we are able to identify clear differences between synthetic vs. real data distributions, our analysis of the errors made by models trained on 0\% vs. 100\% synthetic data failed to yield any generalizable insights that could inform modelling approaches. A more fine-grained study of the effect of using synthetic data on the behaviour of the downstream model is hence left as a subject of future research.

\section{Ethical Considerations}
\label{sec:ethical}
The usage of synthetic data has several important ethical considerations. In the era of LMs trained on internet-wide corpora having poor documentation as to their exact data sources, it becomes challenging to ensure the privacy of individuals whose data may be obtainable via a public crawl~\citep{yao2024survey}. Additionally, models trained on massive internet-based data sources may contain implicit biases, as well as illegal and/or highly offensive material that is hard to audit and clean~\citep{bender2021dangers}. This data affects the synthetic data obtained from prompt models, and could unknowingly impose cultural or ethical viewpoints that are unintended or not well aligned with the use case in mind. Specifically, prior work has shown that one of the prompt models studied in this work, GPT-3.5, often disagrees with humans on key ethical questions~\citep{felkner2024gpt}. The endeavour to completely replace human annotation with synthetic data generation also has key implications on the extent to which the field of NLP employs human annotators. It is possible that an increasing reliance on purely synthetic data reduces the demand for human annotation, which would place a downward pressure on the working standards and compensation awarded to the remaining human annotators~\citep{weidinger2022taxonomy}. We argue in this work that we should not try to eliminate human annotation from our dataset and method design, showing that their work contributes uniquely helpful data points. 

\section{Acknowledgements}
This work was funded by the Defense Advanced Research Projects Agency with award HR00112220046. Any opinions, findings, conclusions, or recommendations expressed here are those of the authors and do not necessarily reflect the view of our sponsors.

This work used Jetstream2 at Indiana University through allocation CIS240665 from the Advanced Cyberinfrastructure Coordination Ecosystem: Services \& Support (ACCESS) program, which is supported by U.S. National Science Foundation grants \#2138259, \#2138286, \#2138307, \#2137603, and \#2138296.


\bibliography{main}

\begin{thebibliography}{64}
\providecommand{\natexlab}[1]{#1}

\bibitem[{Achiam et~al.(2023)Achiam, Adler, Agarwal, Ahmad, Akkaya, Aleman, Almeida, Altenschmidt, Altman, Anadkat et~al.}]{achiam2023gpt}
Josh Achiam, Steven Adler, Sandhini Agarwal, Lama Ahmad, Ilge Akkaya, Florencia~Leoni Aleman, Diogo Almeida, Janko Altenschmidt, Sam Altman, Shyamal Anadkat, et~al. 2023.
\newblock Gpt-4 technical report.
\newblock \emph{arXiv preprint arXiv:2303.08774}.

\bibitem[{Ahmed et~al.(2024)Ahmed, Devanbu, Treude, and Pradel}]{Ahmed2024CanLR}
Toufique Ahmed, Prem Devanbu, Christoph Treude, and Michael Pradel. 2024.
\newblock \href {https://api.semanticscholar.org/CorpusID:271855028} {Can llms replace manual annotation of software engineering artifacts?}
\newblock \emph{ArXiv}, abs/2408.05534.

\bibitem[{Ashok et~al.(2023)Ashok, Kulkarni, Pham, and Poczos}]{ashok2023scifix}
Dhananjay Ashok, Atharva Kulkarni, Hai Pham, and Barnabas Poczos. 2023.
\newblock \href {https://doi.org/10.18653/v1/2023.findings-emnlp.451} {The student becomes the master: Outperforming {GPT}3 on scientific factual error correction}.
\newblock In \emph{Findings of the Association for Computational Linguistics: EMNLP 2023}, pages 6762--6778, Singapore. Association for Computational Linguistics.

\bibitem[{Bender et~al.(2021)Bender, Gebru, McMillan-Major, and Shmitchell}]{bender2021dangers}
Emily~M Bender, Timnit Gebru, Angelina McMillan-Major, and Shmargaret Shmitchell. 2021.
\newblock On the dangers of stochastic parrots: Can language models be too big?
\newblock In \emph{Proceedings of the 2021 ACM conference on fairness, accountability, and transparency}, pages 610--623.

\bibitem[{Bisbee et~al.(2024)Bisbee, Clinton, Dorff, Kenkel, and Larson}]{Bisbee2024SyntheticRF}
James Bisbee, Joshua~D. Clinton, Cassy Dorff, Brenton Kenkel, and Jennifer~M. Larson. 2024.
\newblock \href {https://api.semanticscholar.org/CorpusID:269845858} {Synthetic replacements for human survey data? the perils of large language models}.
\newblock \emph{Political Analysis}.

\bibitem[{Briesch et~al.(2023)Briesch, Sobania, and Rothlauf}]{Briesch2023LargeLM}
Martin Briesch, Dominik Sobania, and Franz Rothlauf. 2023.
\newblock \href {https://api.semanticscholar.org/CorpusID:265466007} {Large language models suffer from their own output: An analysis of the self-consuming training loop}.
\newblock \emph{ArXiv}, abs/2311.16822.

\bibitem[{Brown et~al.(2020)Brown, Mann, Ryder, Subbiah, Kaplan, Dhariwal, Neelakantan, Shyam, Sastry, Askell, Agarwal, Herbert-Voss, Krueger, Henighan, Child, Ramesh, Ziegler, Wu, Winter, Hesse, Chen, Sigler, Litwin, Gray, Chess, Clark, Berner, McCandlish, Radford, Sutskever, and Amodei}]{brown2020language}
Tom Brown, Benjamin Mann, Nick Ryder, Melanie Subbiah, Jared~D Kaplan, Prafulla Dhariwal, Arvind Neelakantan, Pranav Shyam, Girish Sastry, Amanda Askell, Sandhini Agarwal, Ariel Herbert-Voss, Gretchen Krueger, Tom Henighan, Rewon Child, Aditya Ramesh, Daniel Ziegler, Jeffrey Wu, Clemens Winter, Chris Hesse, Mark Chen, Eric Sigler, Mateusz Litwin, Scott Gray, Benjamin Chess, Jack Clark, Christopher Berner, Sam McCandlish, Alec Radford, Ilya Sutskever, and Dario Amodei. 2020.
\newblock \href {https://proceedings.neurips.cc/paper_files/paper/2020/file/1457c0d6bfcb4967418bfb8ac142f64a-Paper.pdf} {Language models are few-shot learners}.
\newblock In \emph{Advances in Neural Information Processing Systems}, volume~33, pages 1877--1901. Curran Associates, Inc.

\bibitem[{Chen et~al.(2024)Chen, Zhang, Wang, Zhao, Wen, and Chen}]{Chen2024UnveilingTF}
Jie Chen, Yupeng Zhang, Bingning Wang, Xin Zhao, Ji-Rong Wen, and Weipeng Chen. 2024.
\newblock \href {https://doi.org/10.18653/v1/2024.findings-emnlp.873} {Unveiling the flaws: Exploring imperfections in synthetic data and mitigation strategies for large language models}.
\newblock In \emph{Findings of the Association for Computational Linguistics: EMNLP 2024}, pages 14855--14865, Miami, Florida, USA. Association for Computational Linguistics.

\bibitem[{Chen et~al.(2012)Chen, Huang, Huang, Liou, and Chang}]{chen2012prefer}
Mei-Hua Chen, Shih-Ting Huang, Chung-Chi Huang, Hsien-Chin Liou, and Jason~S Chang. 2012.
\newblock Prefer: using a graph-based approach to generate paraphrases for language learning.
\newblock In \emph{Proceedings of the Seventh Workshop on Building Educational Applications Using NLP}, pages 80--85.

\bibitem[{De~Gispert et~al.(2005)De~Gispert, Mari{\~n}o, and Crego}]{de2005improving}
Adria De~Gispert, Jos{\'e}~B Mari{\~n}o, and Josep~Maria Crego. 2005.
\newblock Improving statistical machine translation by classifying and generalizing inflected verb forms.
\newblock In \emph{INTERSPEECH}, pages 3193--3196.

\bibitem[{Devlin et~al.(2019)Devlin, Chang, Lee, and Toutanova}]{devlin-etal-2019-bert}
Jacob Devlin, Ming-Wei Chang, Kenton Lee, and Kristina Toutanova. 2019.
\newblock \href {https://doi.org/10.18653/v1/N19-1423} {{BERT}: Pre-training of deep bidirectional transformers for language understanding}.
\newblock In \emph{Proceedings of the 2019 Conference of the North {A}merican Chapter of the Association for Computational Linguistics: Human Language Technologies, Volume 1 (Long and Short Papers)}, pages 4171--4186, Minneapolis, Minnesota. Association for Computational Linguistics.

\bibitem[{Dubey et~al.(2024)Dubey, Jauhri, Pandey, Kadian, Al-Dahle, Letman, Mathur, Schelten, Yang, Fan et~al.}]{dubey2024llama}
Abhimanyu Dubey, Abhinav Jauhri, Abhinav Pandey, Abhishek Kadian, Ahmad Al-Dahle, Aiesha Letman, Akhil Mathur, Alan Schelten, Amy Yang, Angela Fan, et~al. 2024.
\newblock The llama 3 herd of models.
\newblock \emph{arXiv preprint arXiv:2407.21783}.

\bibitem[{Dubois et~al.(2024)Dubois, Li, Taori, Zhang, Gulrajani, Ba, Guestrin, Liang, and Hashimoto}]{Dubois2023AlpacaFarmAS}
Yann Dubois, Chen~Xuechen Li, Rohan Taori, Tianyi Zhang, Ishaan Gulrajani, Jimmy Ba, Carlos Guestrin, Percy~S Liang, and Tatsunori~B Hashimoto. 2024.
\newblock Alpacafarm: A simulation framework for methods that learn from human feedback.
\newblock \emph{Advances in Neural Information Processing Systems}, 36.

\bibitem[{Fan et~al.(2024)Fan, Chen, Krishnan, Katabi, Isola, and Tian}]{Fan2023ScalingLO}
Lijie Fan, Kaifeng Chen, Dilip Krishnan, Dina Katabi, Phillip Isola, and Yonglong Tian. 2024.
\newblock Scaling laws of synthetic images for model training... for now.
\newblock In \emph{Proceedings of the IEEE/CVF Conference on Computer Vision and Pattern Recognition}, pages 7382--7392.

\bibitem[{Felkner et~al.(2024)Felkner, Thompson, and May}]{felkner2024gpt}
Virginia Felkner, Jennifer Thompson, and Jonathan May. 2024.
\newblock \href {https://doi.org/10.18653/v1/2024.acl-long.760} {{GPT} is not an annotator: The necessity of human annotation in fairness benchmark construction}.
\newblock In \emph{Proceedings of the 62nd Annual Meeting of the Association for Computational Linguistics (Volume 1: Long Papers)}, pages 14104--14115, Bangkok, Thailand. Association for Computational Linguistics.

\bibitem[{Gao et~al.(2023)Gao, Xiong, Gao, Jia, Pan, Bi, Dai, Sun, Guo, Wang, and Wang}]{Gao2023RetrievalAugmentedGF}
Yunfan Gao, Yun Xiong, Xinyu Gao, Kangxiang Jia, Jinliu Pan, Yuxi Bi, Yi~Dai, Jiawei Sun, Qianyu Guo, Meng Wang, and Haofen Wang. 2023.
\newblock \href {https://api.semanticscholar.org/CorpusID:266359151} {Retrieval-augmented generation for large language models: A survey}.
\newblock \emph{ArXiv}, abs/2312.10997.

\bibitem[{Guo et~al.(2024)Guo, Shang, Vazirgiannis, and Clavel}]{Guo2023TheCD}
Yanzhu Guo, Guokan Shang, Michalis Vazirgiannis, and Chlo{\'e} Clavel. 2024.
\newblock \href {https://doi.org/10.18653/v1/2024.findings-naacl.228} {The curious decline of linguistic diversity: Training language models on synthetic text}.
\newblock In \emph{Findings of the Association for Computational Linguistics: NAACL 2024}, pages 3589--3604, Mexico City, Mexico. Association for Computational Linguistics.

\bibitem[{Hammoud et~al.(2024)Hammoud, Itani, Pizzati, Torr, Bibi, and Ghanem}]{Hammoud2024SynthCLIPAW}
Hasan Hammoud, Hani Itani, Fabio Pizzati, Philip H.~S. Torr, Adel Bibi, and Bernard Ghanem. 2024.
\newblock \href {https://api.semanticscholar.org/CorpusID:267411953} {Synthclip: Are we ready for a fully synthetic clip training?}
\newblock \emph{ArXiv}, abs/2402.01832.

\bibitem[{He et~al.(2023)He, Sun, Yu, Xue, Zhang, Torr, Bai, and Qi}]{He2022IsSD}
Ruifei He, Shuyang Sun, Xin Yu, Chuhui Xue, Wenqing Zhang, Philip Torr, Song Bai, and Xiaojuan Qi. 2023.
\newblock \href {https://openreview.net/forum?id=nUmCcZ5RKF} {Is synthetic data from generative models ready for image recognition?}
\newblock In \emph{The Eleventh International Conference on Learning Representations}.

\bibitem[{Honovich et~al.(2022)Honovich, Aharoni, Herzig, Taitelbaum, Kukliansy, Cohen, Scialom, Szpektor, Hassidim, and Matias}]{Honovich2022TRUERF}
Or~Honovich, Roee Aharoni, Jonathan Herzig, Hagai Taitelbaum, Doron Kukliansy, Vered Cohen, Thomas Scialom, Idan Szpektor, Avinatan Hassidim, and Yossi Matias. 2022.
\newblock \href {https://doi.org/10.18653/v1/2022.naacl-main.287} {{TRUE}: Re-evaluating factual consistency evaluation}.
\newblock In \emph{Proceedings of the 2022 Conference of the North American Chapter of the Association for Computational Linguistics: Human Language Technologies}, pages 3905--3920, Seattle, United States. Association for Computational Linguistics.

\bibitem[{Hosseini et~al.(2024)Hosseini, Petrov, Fabrikant, and Louis}]{Hosseini2024ASD}
Mohammad~Javad Hosseini, Andrey Petrov, Alex Fabrikant, and Annie Louis. 2024.
\newblock \href {https://doi.org/10.18653/v1/2024.acl-long.120} {A synthetic data approach for domain generalization of {NLI} models}.
\newblock In \emph{Proceedings of the 62nd Annual Meeting of the Association for Computational Linguistics (Volume 1: Long Papers)}, pages 2212--2226, Bangkok, Thailand. Association for Computational Linguistics.

\bibitem[{Hu et~al.(2022)Hu, yelong shen, Wallis, Allen-Zhu, Li, Wang, Wang, and Chen}]{hu2021lora}
Edward~J Hu, yelong shen, Phillip Wallis, Zeyuan Allen-Zhu, Yuanzhi Li, Shean Wang, Lu~Wang, and Weizhu Chen. 2022.
\newblock \href {https://openreview.net/forum?id=nZeVKeeFYf9} {Lo{RA}: Low-rank adaptation of large language models}.
\newblock In \emph{International Conference on Learning Representations}.

\bibitem[{Kobayashi(2018)}]{Kobayashi2018ContextualAD}
Sosuke Kobayashi. 2018.
\newblock \href {https://doi.org/10.18653/v1/N18-2072} {Contextual augmentation: Data augmentation by words with paradigmatic relations}.
\newblock In \emph{Proceedings of the 2018 Conference of the North {A}merican Chapter of the Association for Computational Linguistics: Human Language Technologies, Volume 2 (Short Papers)}, pages 452--457, New Orleans, Louisiana. Association for Computational Linguistics.

\bibitem[{Krizhevsky et~al.(2012)Krizhevsky, Sutskever, and Hinton}]{krizhevsky2012imagenet}
Alex Krizhevsky, Ilya Sutskever, and Geoffrey~E Hinton. 2012.
\newblock Imagenet classification with deep convolutional neural networks.
\newblock \emph{Advances in neural information processing systems}, 25.

\bibitem[{Li et~al.(2024)Li, Yu, Zhou, Schick, Levy, Zettlemoyer, Weston, and Lewis}]{Li2023SelfAlignmentWI}
Xian Li, Ping Yu, Chunting Zhou, Timo Schick, Omer Levy, Luke Zettlemoyer, Jason~E Weston, and Mike Lewis. 2024.
\newblock \href {https://openreview.net/forum?id=1oijHJBRsT} {Self-alignment with instruction backtranslation}.
\newblock In \emph{The Twelfth International Conference on Learning Representations}.

\bibitem[{Li et~al.(2023)Li, Zhu, Lu, and Yin}]{Li2023SyntheticDG}
Zhuoyan Li, Hangxiao Zhu, Zhuoran Lu, and Ming Yin. 2023.
\newblock \href {https://openreview.net/forum?id=MmBjKmHIND} {Synthetic data generation with large language models for text classification: Potential and limitations}.
\newblock In \emph{The 2023 Conference on Empirical Methods in Natural Language Processing}.

\bibitem[{Lin(2004)}]{lin-2004-rouge}
Chin-Yew Lin. 2004.
\newblock \href {https://aclanthology.org/W04-1013} {{ROUGE}: A package for automatic evaluation of summaries}.
\newblock In \emph{Text Summarization Branches Out}, pages 74--81, Barcelona, Spain. Association for Computational Linguistics.

\bibitem[{Lin et~al.(2019)Lin, Tafjord, Clark, and Gardner}]{Lin2019ReasoningOP}
Kevin Lin, Oyvind Tafjord, Peter Clark, and Matt Gardner. 2019.
\newblock \href {https://doi.org/10.18653/v1/D19-5808} {Reasoning over paragraph effects in situations}.
\newblock In \emph{Proceedings of the 2nd Workshop on Machine Reading for Question Answering}, pages 58--62, Hong Kong, China. Association for Computational Linguistics.

\bibitem[{Liu et~al.(2022)Liu, Swayamdipta, Smith, and Choi}]{Liu2022WANLIWA}
Alisa Liu, Swabha Swayamdipta, Noah~A. Smith, and Yejin Choi. 2022.
\newblock \href {https://api.semanticscholar.org/CorpusID:246016339} {Wanli: Worker and ai collaboration for natural language inference dataset creation}.
\newblock In \emph{Conference on Empirical Methods in Natural Language Processing}.

\bibitem[{Lo et~al.(2020)Lo, Wang, Neumann, Kinney, and Weld}]{lo2019s2orc}
Kyle Lo, Lucy~Lu Wang, Mark Neumann, Rodney Kinney, and Daniel Weld. 2020.
\newblock \href {https://doi.org/10.18653/v1/2020.acl-main.447} {{S}2{ORC}: The semantic scholar open research corpus}.
\newblock In \emph{Proceedings of the 58th Annual Meeting of the Association for Computational Linguistics}, pages 4969--4983, Online. Association for Computational Linguistics.

\bibitem[{Meng et~al.(2022)Meng, Huang, Zhang, and Han}]{Meng2022GeneratingTD}
Yu~Meng, Jiaxin Huang, Yu~Zhang, and Jiawei Han. 2022.
\newblock Generating training data with language models: Towards zero-shot language understanding.
\newblock \emph{Advances in Neural Information Processing Systems}, 35:462--477.

\bibitem[{Mishra et~al.(2022)Mishra, Suryavardan, Bhaskar, Chopra, Reganti, Patwa, Das, Chakraborty, Sheth, and Ekbal}]{Mishra2022FACTIFYAM}
Shreyash Mishra, S~Suryavardan, Amrit Bhaskar, Parul Chopra, Aishwarya~N. Reganti, Parth Patwa, Amitava Das, Tanmoy Chakraborty, A.~Sheth, and Asif Ekbal. 2022.
\newblock \href {https://api.semanticscholar.org/CorpusID:252016186} {Factify: A multi-modal fact verification dataset}.
\newblock In \emph{DE-FACTIFY@AAAI}.

\bibitem[{Ni et~al.(2024)Ni, Shi, Stammbach, Sachan, Ash, and Leippold}]{Ni2024AFaCTAAT}
Jingwei Ni, Minjing Shi, Dominik Stammbach, Mrinmaya Sachan, Elliott Ash, and Markus Leippold. 2024.
\newblock \href {https://api.semanticscholar.org/CorpusID:267750827} {Afacta: Assisting the annotation of factual claim detection with reliable llm annotators}.
\newblock In \emph{Annual Meeting of the Association for Computational Linguistics}.

\bibitem[{Papineni et~al.(2002)Papineni, Roukos, Ward, and Zhu}]{papineni-etal-2002-bleu}
Kishore Papineni, Salim Roukos, Todd Ward, and Wei-Jing Zhu. 2002.
\newblock \href {https://doi.org/10.3115/1073083.1073135} {{B}leu: a method for automatic evaluation of machine translation}.
\newblock In \emph{Proceedings of the 40th Annual Meeting of the Association for Computational Linguistics}, pages 311--318, Philadelphia, Pennsylvania, USA. Association for Computational Linguistics.

\bibitem[{Reddy et~al.(2019)Reddy, Chen, and Manning}]{reddy2019coqa}
Siva Reddy, Danqi Chen, and Christopher~D. Manning. 2019.
\newblock \href {https://doi.org/10.1162/tacl_a_00266} {{C}o{QA}: A conversational question answering challenge}.
\newblock \emph{Transactions of the Association for Computational Linguistics}, 7:249--266.

\bibitem[{Schimanski et~al.(2024)Schimanski, Ni, Kraus, Ash, and Leippold}]{Schimanski2024TowardsFA}
Tobias Schimanski, Jingwei Ni, Mathias Kraus, Elliott Ash, and Markus Leippold. 2024.
\newblock \href {https://doi.org/10.18653/v1/2024.acl-long.105} {Towards faithful and robust {LLM} specialists for evidence-based question-answering}.
\newblock In \emph{Proceedings of the 62nd Annual Meeting of the Association for Computational Linguistics (Volume 1: Long Papers)}, pages 1913--1931, Bangkok, Thailand. Association for Computational Linguistics.

\bibitem[{Seddik et~al.(2024)Seddik, Chen, Hayou, Youssef, and Debbah}]{Seddik2024HowBI}
Mohamed El~Amine Seddik, Suei-Wen Chen, Soufiane Hayou, Pierre Youssef, and M{\'e}rouane Debbah. 2024.
\newblock \href {https://api.semanticscholar.org/CorpusID:269005923} {How bad is training on synthetic data? a statistical analysis of language model collapse}.
\newblock \emph{ArXiv}, abs/2404.05090.

\bibitem[{Sennrich et~al.(2016)Sennrich, Haddow, and Birch}]{Sennrich2015ImprovingNM}
Rico Sennrich, Barry Haddow, and Alexandra Birch. 2016.
\newblock \href {https://doi.org/10.18653/v1/P16-1009} {Improving neural machine translation models with monolingual data}.
\newblock In \emph{Proceedings of the 54th Annual Meeting of the Association for Computational Linguistics (Volume 1: Long Papers)}, pages 86--96, Berlin, Germany. Association for Computational Linguistics.

\bibitem[{Sharma et~al.(2023)Sharma, Saran, and Patil}]{Sharma2023FakeND}
Umang Sharma, Sidarth Saran, and Dr~Shankar~M. Patil. 2023.
\newblock \href {https://api.semanticscholar.org/CorpusID:235387137} {Fake news detection using machine learning algorithms}.
\newblock \emph{2023 International Conference on New Frontiers in Communication, Automation, Management and Security (ICCAMS)}, 1:1--7.

\bibitem[{Shumailov et~al.(2023)Shumailov, Shumaylov, Zhao, Gal, Papernot, and Anderson}]{Shumailov2023TheCO}
Ilia Shumailov, Zakhar Shumaylov, Yiren Zhao, Yarin Gal, Nicolas Papernot, and Ross Anderson. 2023.
\newblock \href {https://api.semanticscholar.org/CorpusID:258987240} {The curse of recursion: Training on generated data makes models forget}.
\newblock \emph{ArXiv}, abs/2305.17493.

\bibitem[{Simard et~al.(2002)Simard, LeCun, Denker, and Victorri}]{simard2002transformation}
Patrice~Y Simard, Yann~A LeCun, John~S Denker, and Bernard Victorri. 2002.
\newblock Transformation invariance in pattern recognition—tangent distance and tangent propagation.
\newblock In \emph{Neural networks: tricks of the trade}, pages 239--274. Springer.

\bibitem[{Singh et~al.(2024)Singh, Navaratnam, Holmer, Schaub-Meyer, and Roth}]{Singh2024IsSD}
Krishnakant Singh, Thanush Navaratnam, Jannik Holmer, Simone Schaub-Meyer, and Stefan Roth. 2024.
\newblock Is synthetic data all we need? benchmarking the robustness of models trained with synthetic images.
\newblock In \emph{Proceedings of the IEEE/CVF Conference on Computer Vision and Pattern Recognition}, pages 2505--2515.

\bibitem[{Tan et~al.(2024)Tan, Li, Wang, Beigi, Jiang, Bhattacharjee, Karami, Li, Cheng, and Liu}]{Tan2024LargeLM}
Zhen Tan, Dawei Li, Song Wang, Alimohammad Beigi, Bohan Jiang, Amrita Bhattacharjee, Mansooreh Karami, Jundong Li, Lu~Cheng, and Huan Liu. 2024.
\newblock \href {https://doi.org/10.18653/v1/2024.emnlp-main.54} {Large language models for data annotation and synthesis: A survey}.
\newblock In \emph{Proceedings of the 2024 Conference on Empirical Methods in Natural Language Processing}, pages 930--957, Miami, Florida, USA. Association for Computational Linguistics.

\bibitem[{Tang et~al.(2024)Tang, Laban, and Durrett}]{Tang2024MiniCheckEF}
Liyan Tang, Philippe Laban, and Greg Durrett. 2024.
\newblock \href {https://api.semanticscholar.org/CorpusID:269157443} {Minicheck: Efficient fact-checking of llms on grounding documents}.
\newblock \emph{ArXiv}, abs/2404.10774.

\bibitem[{Tang et~al.(2023)Tang, Han, Jiang, and Hu}]{Tang2023DoesSD}
Ruixiang Tang, Xiaotian Han, Xiaoqian Jiang, and Xia Hu. 2023.
\newblock \href {https://api.semanticscholar.org/CorpusID:257405132} {Does synthetic data generation of llms help clinical text mining?}
\newblock \emph{ArXiv}, abs/2303.04360.

\bibitem[{Thorne et~al.(2018)Thorne, Vlachos, Christodoulopoulos, and Mittal}]{thorne2018fever}
James Thorne, Andreas Vlachos, Christos Christodoulopoulos, and Arpit Mittal. 2018.
\newblock Fever: a large-scale dataset for fact extraction and verification.
\newblock In \emph{Proceedings of the 2018 Conference of the North American Chapter of the Association for Computational Linguistics: Human Language Technologies, Volume 1 (Long Papers)}, pages 809--819.

\bibitem[{Villalobos et~al.(2024)Villalobos, Ho, Sevilla, Besiroglu, Heim, and Hobbhahn}]{Villalobos2022WillWR}
Pablo Villalobos, Anson Ho, Jaime Sevilla, Tamay Besiroglu, Lennart Heim, and Marius Hobbhahn. 2024.
\newblock \href {https://openreview.net/forum?id=ViZcgDQjyG} {Position: Will we run out of data? limits of {LLM} scaling based on human-generated data}.
\newblock In \emph{Forty-first International Conference on Machine Learning}.

\bibitem[{Wadden et~al.(2020)Wadden, Lin, Lo, Wang, van Zuylen, Cohan, and Hajishirzi}]{wadden2020fact}
David Wadden, Shanchuan Lin, Kyle Lo, Lucy~Lu Wang, Madeleine van Zuylen, Arman Cohan, and Hannaneh Hajishirzi. 2020.
\newblock \href {https://doi.org/10.18653/v1/2020.emnlp-main.609} {Fact or fiction: Verifying scientific claims}.
\newblock In \emph{Proceedings of the 2020 Conference on Empirical Methods in Natural Language Processing (EMNLP)}, pages 7534--7550, Online. Association for Computational Linguistics.

\bibitem[{Wan et~al.(2024)Wan, Liu, Ajith, Grazian, Hoex, Zhang, Kit, Xie, and Foster}]{Wan2024SciQAGAF}
Yuwei Wan, Yixuan Liu, Aswathy Ajith, Clara Grazian, Bram Hoex, Wenjie Zhang, Chunyu Kit, Tong Xie, and Ian Foster. 2024.
\newblock \href {https://api.semanticscholar.org/CorpusID:269791295} {Sciqag: A framework for auto-generated science question answering dataset with fine-grained evaluation}.

\bibitem[{Wang et~al.(2021)Wang, Liu, Xu, Zhu, and Zeng}]{wang-etal-2021-want-reduce}
Shuohang Wang, Yang Liu, Yichong Xu, Chenguang Zhu, and Michael Zeng. 2021.
\newblock \href {https://doi.org/10.18653/v1/2021.findings-emnlp.354} {Want to reduce labeling cost? {GPT}-3 can help}.
\newblock In \emph{Findings of the Association for Computational Linguistics: EMNLP 2021}, pages 4195--4205, Punta Cana, Dominican Republic. Association for Computational Linguistics.

\bibitem[{Wang and Yang(2015)}]{Wang2015ThatsSA}
William~Yang Wang and Diyi Yang. 2015.
\newblock \href {https://doi.org/10.18653/v1/D15-1306} {That{'}s so annoying!!!: A lexical and frame-semantic embedding based data augmentation approach to automatic categorization of annoying behaviors using {\#}petpeeve tweets}.
\newblock In \emph{Proceedings of the 2015 Conference on Empirical Methods in Natural Language Processing}, pages 2557--2563, Lisbon, Portugal. Association for Computational Linguistics.

\bibitem[{Wang et~al.(2018)Wang, Pham, Dai, and Neubig}]{Wang2018SwitchOutAE}
Xinyi Wang, Hieu Pham, Zihang Dai, and Graham Neubig. 2018.
\newblock \href {https://doi.org/10.18653/v1/D18-1100} {{S}witch{O}ut: an efficient data augmentation algorithm for neural machine translation}.
\newblock In \emph{Proceedings of the 2018 Conference on Empirical Methods in Natural Language Processing}, pages 856--861, Brussels, Belgium. Association for Computational Linguistics.

\bibitem[{Wei et~al.(2022)Wei, Tay, Bommasani, Raffel, Zoph, Borgeaud, Yogatama, Bosma, Zhou, Metzler et~al.}]{wei2022emergent}
Jason Wei, Yi~Tay, Rishi Bommasani, Colin Raffel, Barret Zoph, Sebastian Borgeaud, Dani Yogatama, Maarten Bosma, Denny Zhou, Donald Metzler, et~al. 2022.
\newblock Emergent abilities of large language models.
\newblock \emph{arXiv preprint arXiv:2206.07682}.

\bibitem[{Weidinger et~al.(2022)Weidinger, Uesato, Rauh, Griffin, Huang, Mellor, Glaese, Cheng, Balle, Kasirzadeh et~al.}]{weidinger2022taxonomy}
Laura Weidinger, Jonathan Uesato, Maribeth Rauh, Conor Griffin, Po-Sen Huang, John Mellor, Amelia Glaese, Myra Cheng, Borja Balle, Atoosa Kasirzadeh, et~al. 2022.
\newblock Taxonomy of risks posed by language models.
\newblock In \emph{Proceedings of the 2022 ACM Conference on Fairness, Accountability, and Transparency}, pages 214--229.

\bibitem[{Weston et~al.(2015)Weston, Bordes, Chopra, and Mikolov}]{Weston2015TowardsAQ}
Jason Weston, Antoine Bordes, Sumit Chopra, and Tomas Mikolov. 2015.
\newblock \href {https://api.semanticscholar.org/CorpusID:3178759} {Towards ai-complete question answering: A set of prerequisite toy tasks}.
\newblock \emph{arXiv: Artificial Intelligence}.

\bibitem[{Williams et~al.(2018)Williams, Nangia, and Bowman}]{MNLI}
Adina Williams, Nikita Nangia, and Samuel Bowman. 2018.
\newblock \href {http://aclweb.org/anthology/N18-1101} {A broad-coverage challenge corpus for sentence understanding through inference}.
\newblock In \emph{Proceedings of the 2018 Conference of the North American Chapter of the Association for Computational Linguistics: Human Language Technologies, Volume 1 (Long Papers)}, pages 1112--1122. Association for Computational Linguistics.

\bibitem[{Wu et~al.(2022)Wu, Madotto, Liu, Fung, and Xiong}]{Wu2021QAConvQA}
Chien-Sheng Wu, Andrea Madotto, Wenhao Liu, Pascale Fung, and Caiming Xiong. 2022.
\newblock \href {https://doi.org/10.18653/v1/2022.acl-long.370} {{QAC}onv: Question answering on informative conversations}.
\newblock In \emph{Proceedings of the 60th Annual Meeting of the Association for Computational Linguistics (Volume 1: Long Papers)}, pages 5389--5411, Dublin, Ireland. Association for Computational Linguistics.

\bibitem[{Xie et~al.(2017)Xie, Wang, Li, L{\'e}vy, Nie, Jurafsky, and Ng}]{Xie2017DataNA}
Ziang Xie, Sida~I. Wang, Jiwei Li, Daniel L{\'e}vy, Aiming Nie, Dan Jurafsky, and Andrew~Y. Ng. 2017.
\newblock \href {https://openreview.net/forum?id=H1VyHY9gg} {Data noising as smoothing in neural network language models}.
\newblock In \emph{International Conference on Learning Representations}.

\bibitem[{Xu et~al.(2022)Xu, Wang, Yu, Ritchie, Yao, Wu, Zhang, Li, Bradford, Sun, Hoang, Sang, Hou, Ma, Yang, Peng, Yu, and Warschauer}]{xu2022fantastic}
Ying Xu, Dakuo Wang, Mo~Yu, Daniel Ritchie, Bingsheng Yao, Tongshuang Wu, Zheng Zhang, Toby Jia-Jun Li, Nora Bradford, Branda Sun, Tran~Bao Hoang, Yisi Sang, Yufang Hou, Xiaojuan Ma, Diyi Yang, Nanyun Peng, Zhou Yu, and Mark Warschauer. 2022.
\newblock \href {https://doi.org/10.18653/v1/2022.acl-long.34} {Fantastic questions and where to find them: {F}airytale{QA} {--} an authentic dataset for narrative comprehension}.
\newblock In \emph{Proceedings of the 60th Annual Meeting of the Association for Computational Linguistics (Volume 1: Long Papers)}, pages 447--460, Dublin, Ireland. Association for Computational Linguistics.

\bibitem[{Yao et~al.(2024)Yao, Duan, Xu, Cai, Sun, and Zhang}]{yao2024survey}
Yifan Yao, Jinhao Duan, Kaidi Xu, Yuanfang Cai, Zhibo Sun, and Yue Zhang. 2024.
\newblock A survey on large language model (llm) security and privacy: The good, the bad, and the ugly.
\newblock \emph{High-Confidence Computing}, page 100211.

\bibitem[{Ye et~al.(2022)Ye, Gao, Li, Xu, Feng, Wu, Yu, and Kong}]{Ye2022ZeroGenEZ}
Jiacheng Ye, Jiahui Gao, Qintong Li, Hang Xu, Jiangtao Feng, Zhiyong Wu, Tao Yu, and Lingpeng Kong. 2022.
\newblock \href {https://doi.org/10.18653/v1/2022.emnlp-main.801} {{Z}ero{G}en: Efficient zero-shot learning via dataset generation}.
\newblock In \emph{Proceedings of the 2022 Conference on Empirical Methods in Natural Language Processing}, pages 11653--11669, Abu Dhabi, United Arab Emirates. Association for Computational Linguistics.

\bibitem[{Yu et~al.(2018)Yu, Dohan, Le, Luong, Zhao, and Chen}]{Yu2018QANetCL}
Adams~Wei Yu, David Dohan, Quoc Le, Thang Luong, Rui Zhao, and Kai Chen. 2018.
\newblock \href {https://openreview.net/forum?id=B14TlG-RW} {Fast and accurate reading comprehension by combining self-attention and convolution}.
\newblock In \emph{International Conference on Learning Representations}.

\bibitem[{Zha et~al.(2023)Zha, Yang, Li, and Hu}]{Zha2023AlignScoreEF}
Yuheng Zha, Yichi Yang, Ruichen Li, and Zhiting Hu. 2023.
\newblock \href {https://doi.org/10.18653/v1/2023.acl-long.634} {{A}lign{S}core: Evaluating factual consistency with a unified alignment function}.
\newblock In \emph{Proceedings of the 61st Annual Meeting of the Association for Computational Linguistics (Volume 1: Long Papers)}, pages 11328--11348, Toronto, Canada. Association for Computational Linguistics.

\bibitem[{Zhang et~al.(2020)Zhang, Kishore, Wu, Weinberger, and Artzi}]{zhang2019bertscore}
Tianyi Zhang, Varsha Kishore, Felix Wu, Kilian~Q. Weinberger, and Yoav Artzi. 2020.
\newblock \href {https://openreview.net/forum?id=SkeHuCVFDr} {Bertscore: Evaluating text generation with bert}.
\newblock In \emph{International Conference on Learning Representations}.

\end{thebibliography}

\clearpage

\appendix
\section{Supplemental Figures}
\label{sec:supplement}
We present a detailed set of figures and tables to supplement the results presented in the main text.

\begin{figure*}
    \centering
    \includegraphics[width=0.7\linewidth]{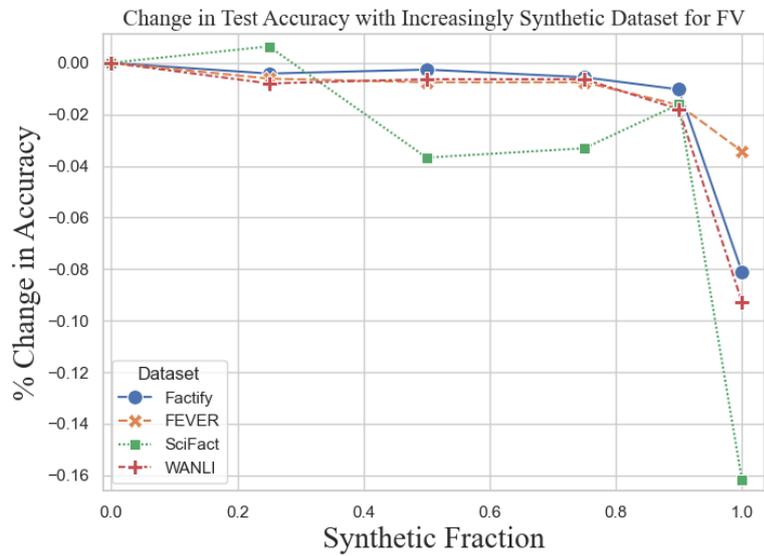}
    
    \vspace{0.02\linewidth}
    \includegraphics[width=0.7\linewidth]{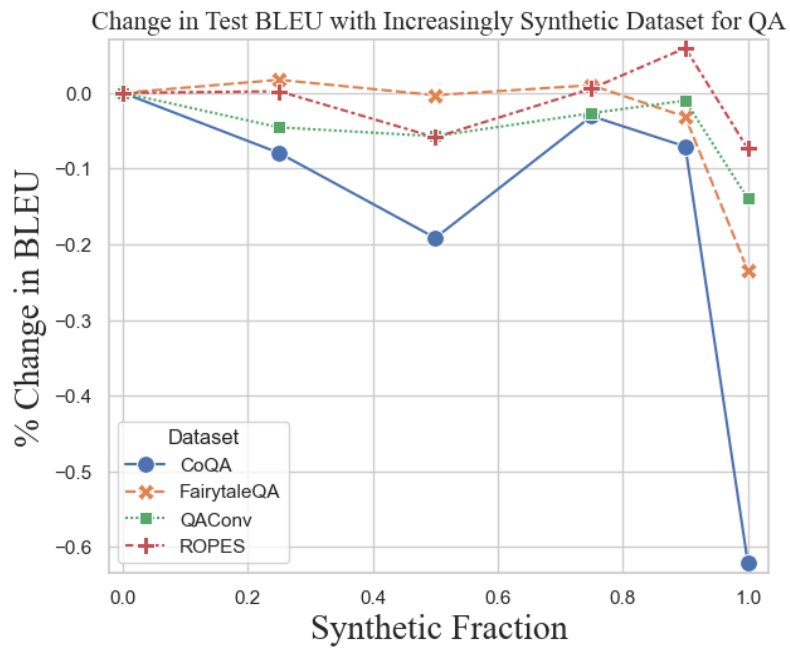}
     \caption{Change in model performance as the proportion of synthetic points in the training data is increased. Across datasets, the performance decrease when moving from 0\% to 90\% synthetic data is often less than that of moving from 90\% to purely synthetic data.}
    \label{fig:fulldataboth}
\end{figure*}

\begin{figure*}
    \centering
    \includegraphics[width=0.7\linewidth]{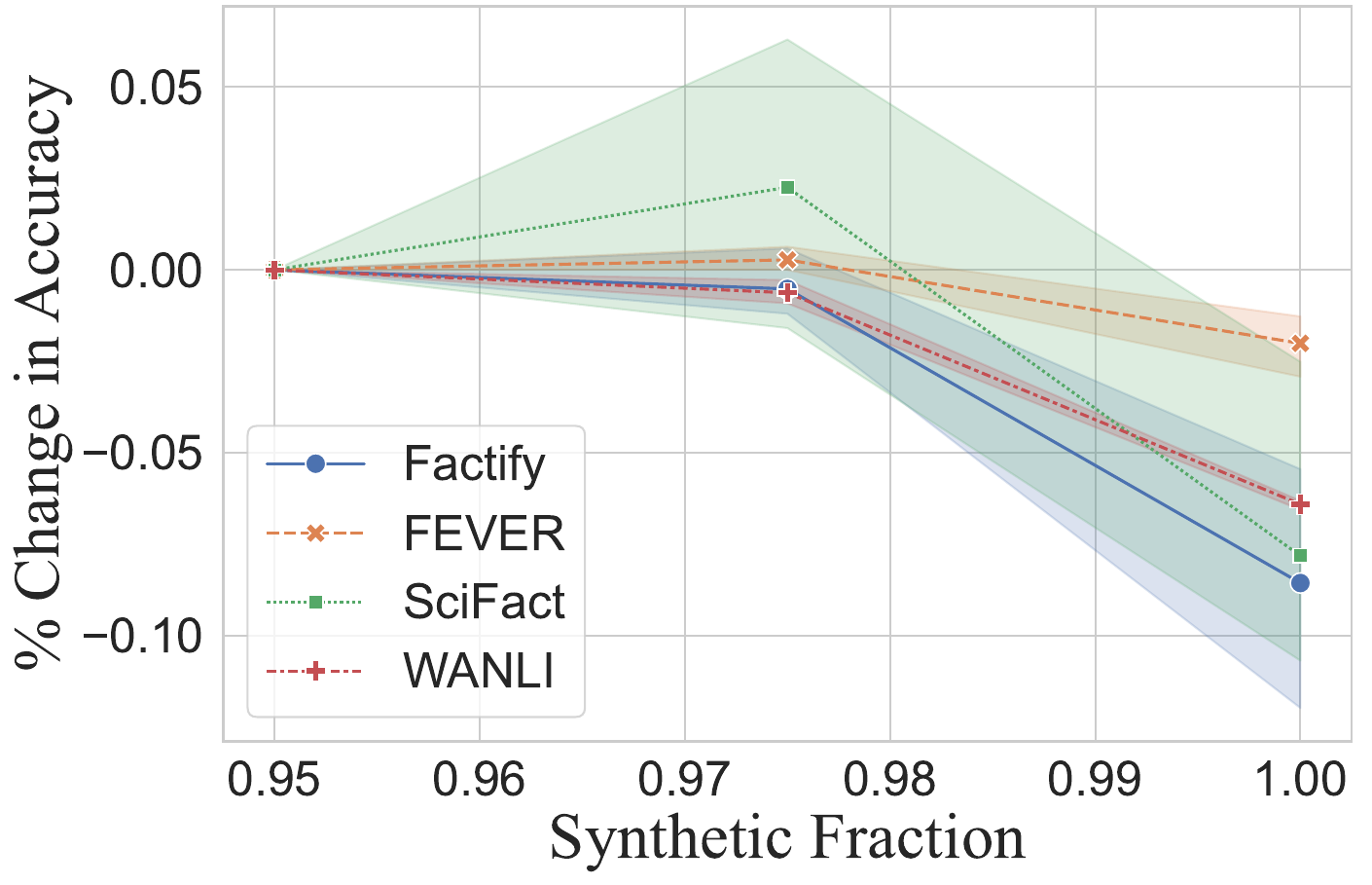}
    
    \vspace{0.02\linewidth}
    \includegraphics[width=0.7\linewidth]{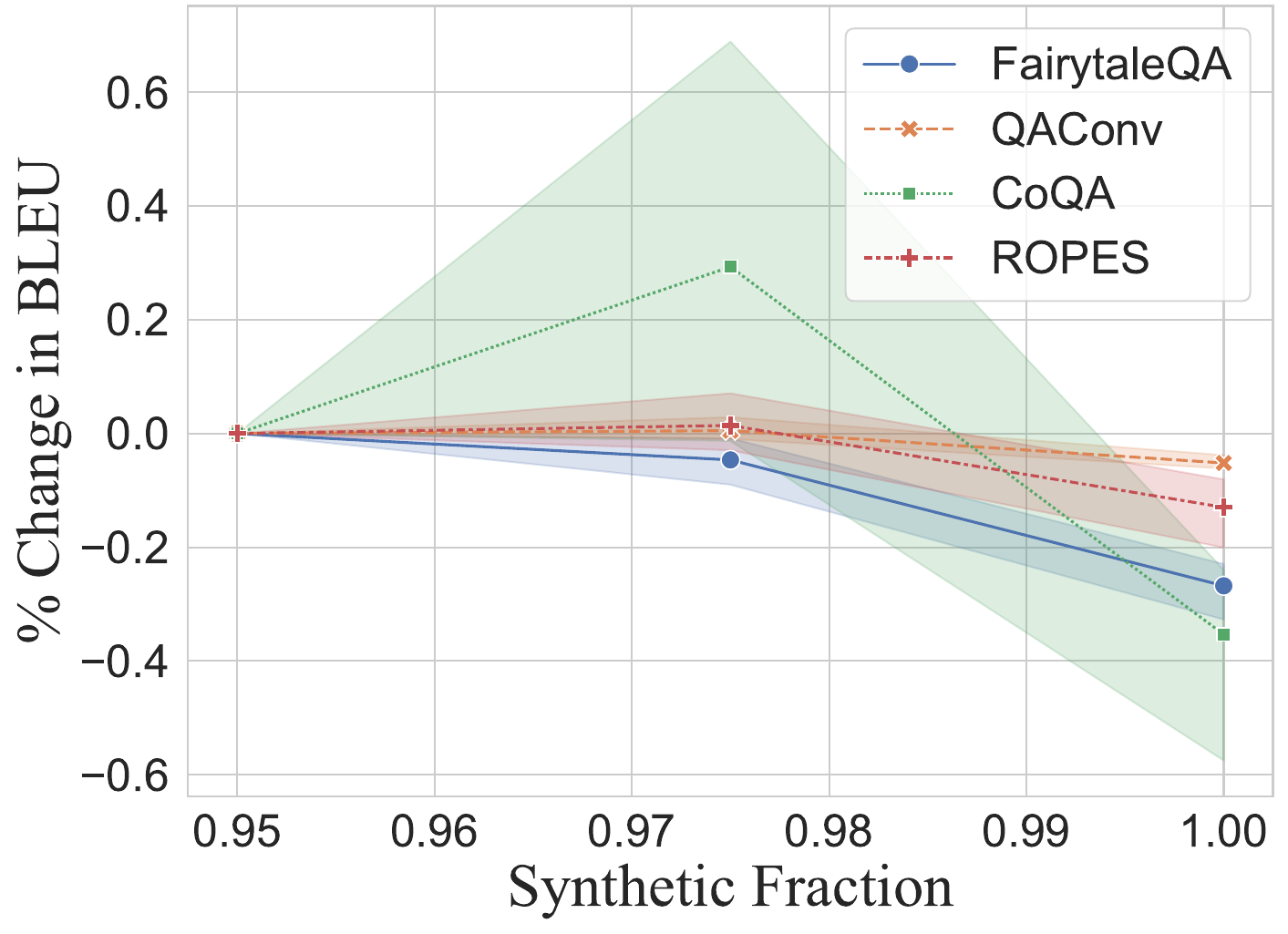}
     \caption{Model performance as the synthetic proportion of the training data varies from 95\% to 100\%. Across all datasets and random seeds, having \textbf{just 2.5\%} of the training dataset being human generated boosts performance.}
    \label{fig:zoom5both}
\end{figure*}

\begin{figure*}[ht]
     \begin{subfigure}[b]{0.4\textwidth}
         \includegraphics[width=\textwidth]{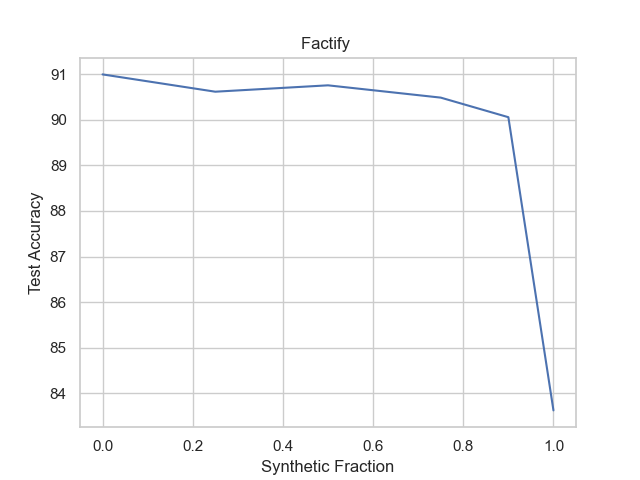}
     \end{subfigure}
     \hfill
     \begin{subfigure}[b]{0.4\textwidth}
         \includegraphics[width=\textwidth]{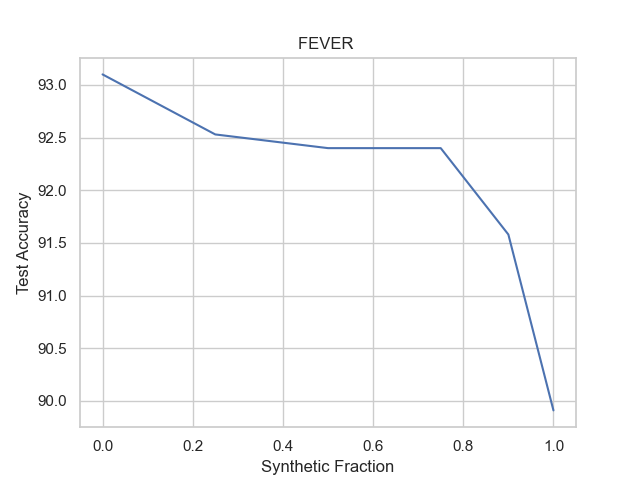}
     \end{subfigure}
     \hfill
     \begin{subfigure}[b]{0.4\textwidth}
         \centering
         \includegraphics[width=\textwidth]{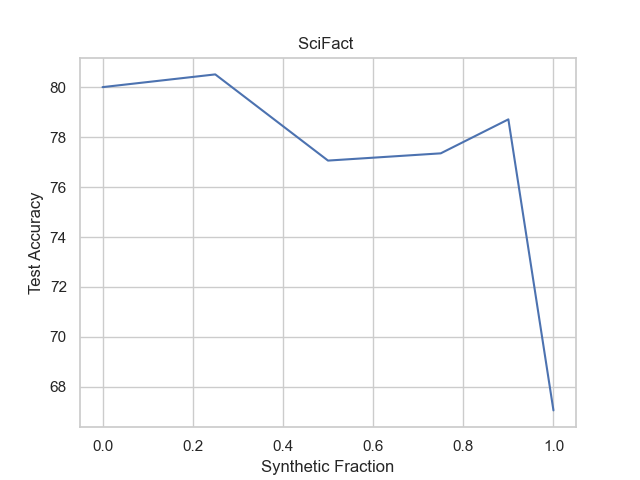}
     \end{subfigure}
     \hfill
     \begin{subfigure}[b]{0.4\textwidth}
         \centering
         \includegraphics[width=\textwidth]{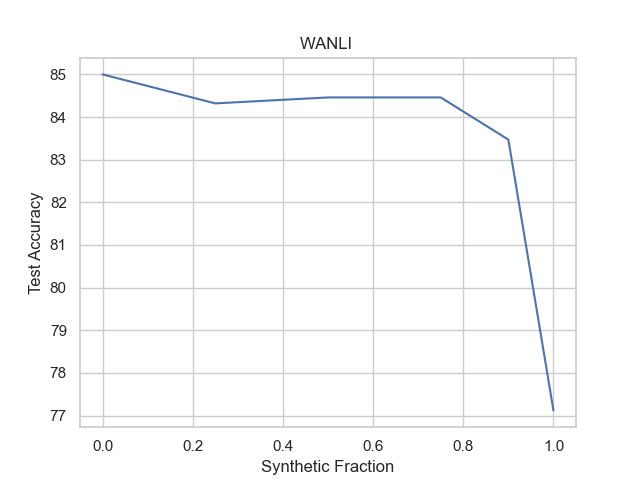}
     \end{subfigure}
      \vfill
     \begin{subfigure}[b]{0.4\textwidth}
         \centering
         \includegraphics[width=\textwidth]{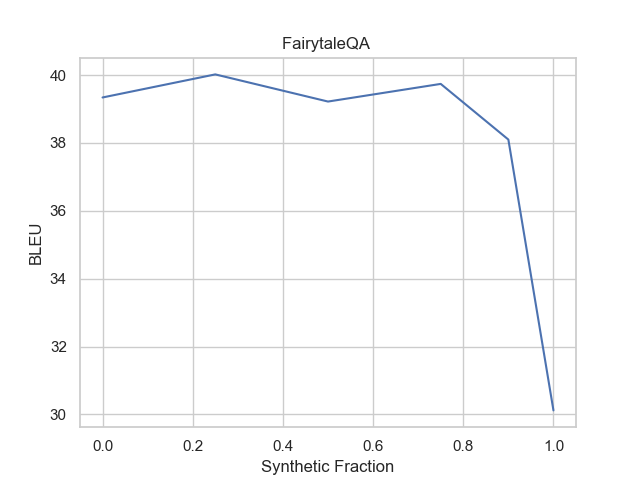}
     \end{subfigure}
     \hfill
     \begin{subfigure}[b]{0.4\textwidth}
         \centering
         \includegraphics[width=\textwidth]{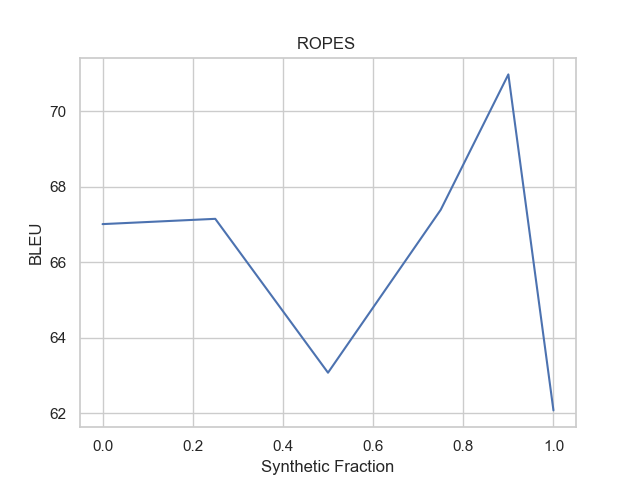}
     \end{subfigure}
     \hfill
     \begin{subfigure}[b]{0.4\textwidth}
         \centering
         \includegraphics[width=\textwidth]{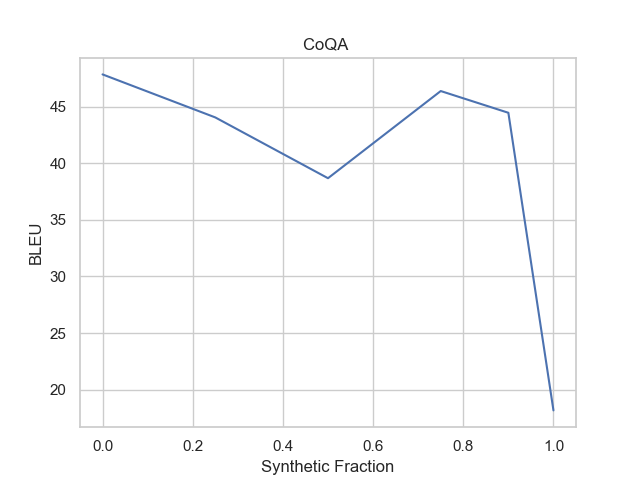}
     \end{subfigure}
     \hfill
     \begin{subfigure}[b]{0.4\textwidth}
         \centering
         \includegraphics[width=\textwidth]{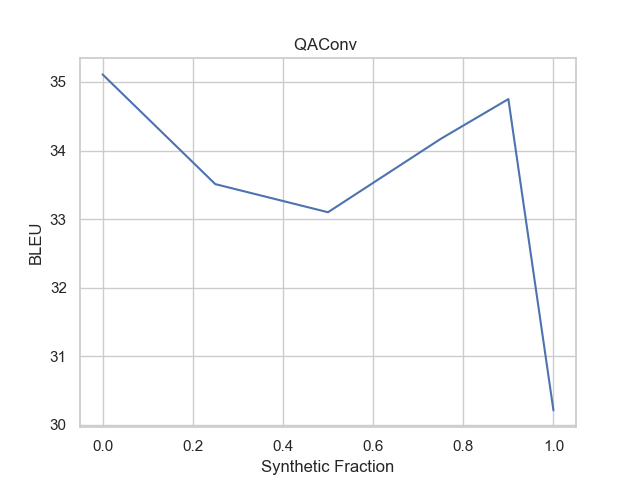}
     \end{subfigure}
        \caption{Change in model performance as the proportion of synthetic points in the training data is varied. Across datasets, the performance decrease when moving from 0\% to 90\% synthetic data is often less than that of moving from 90\% to purely synthetic data.}
        \label{fig:fullind}
\end{figure*}

\begin{figure*}[ht]
     \begin{subfigure}[b]{0.45\textwidth}
         \includegraphics[width=\textwidth]{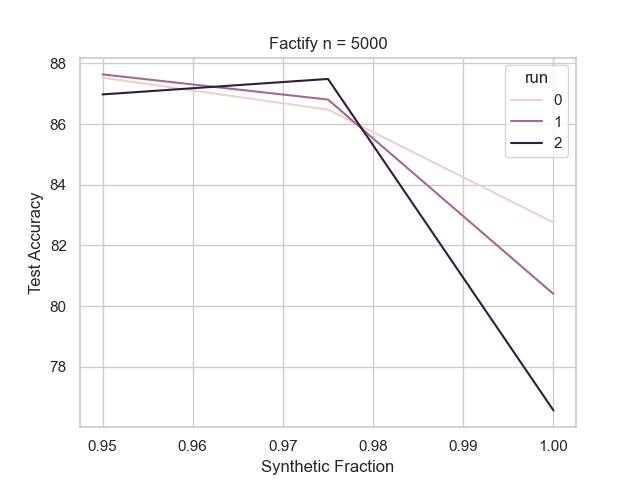}
     \end{subfigure}
     \hfill
     \begin{subfigure}[b]{0.45\textwidth}
         \includegraphics[width=\textwidth]{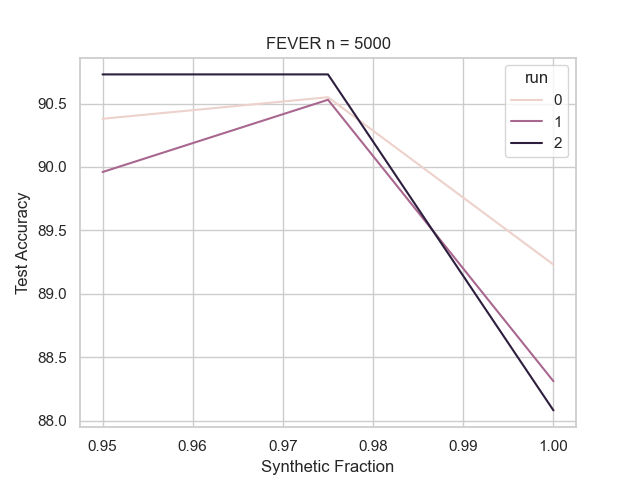}
     \end{subfigure}
     \hfill
     \begin{subfigure}[b]{0.45\textwidth}
         \centering
         \includegraphics[width=\textwidth]{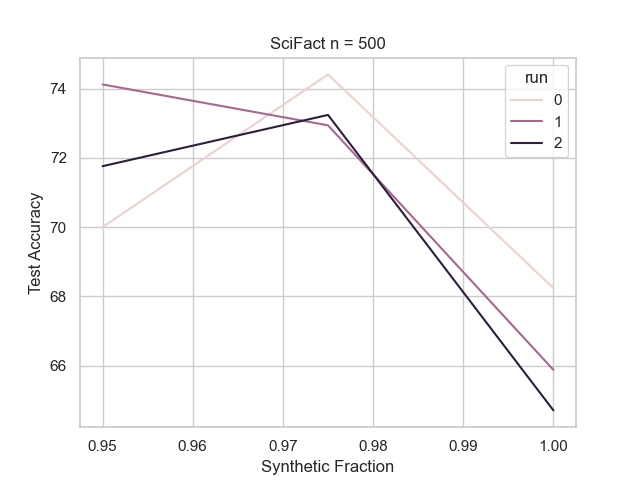}
     \end{subfigure}
     \hfill
     \begin{subfigure}[b]{0.45\textwidth}
         \centering
         \includegraphics[width=\textwidth]{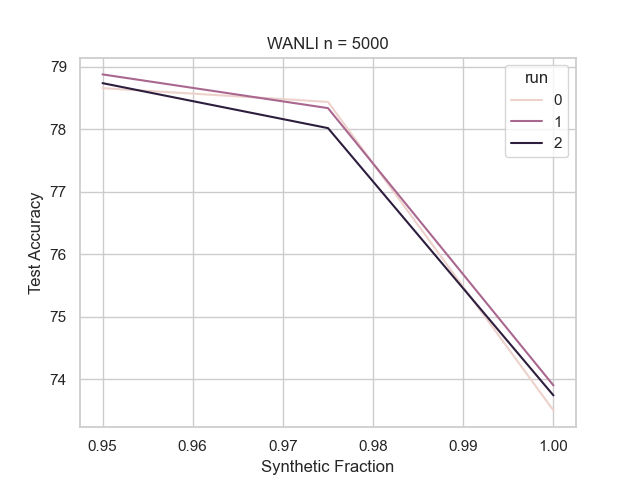}
     \end{subfigure}
      \vfill
     \begin{subfigure}[b]{0.45\textwidth}
         \centering
         \includegraphics[width=\textwidth]{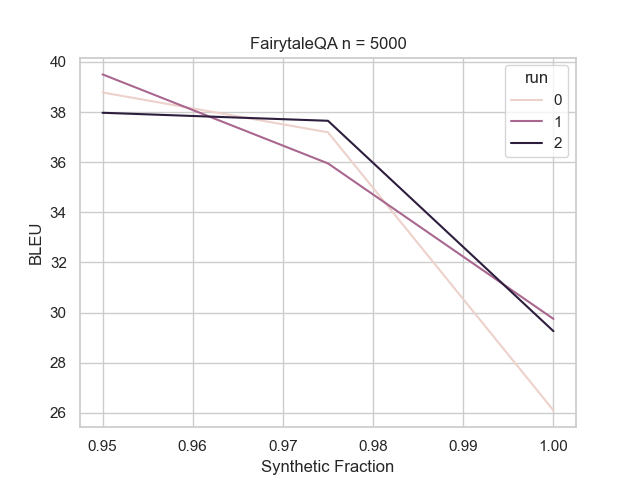}
     \end{subfigure}
     \hfill
     \begin{subfigure}[b]{0.45\textwidth}
         \centering
         \includegraphics[width=\textwidth]{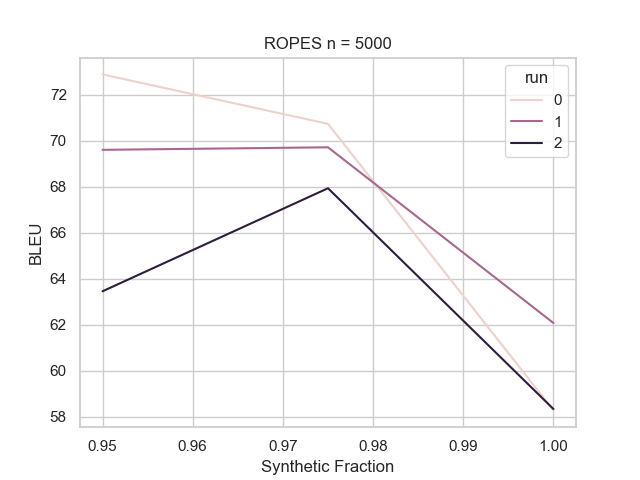}
     \end{subfigure}
     \hfill
     \begin{subfigure}[b]{0.45\textwidth}
         \centering
         \includegraphics[width=\textwidth]{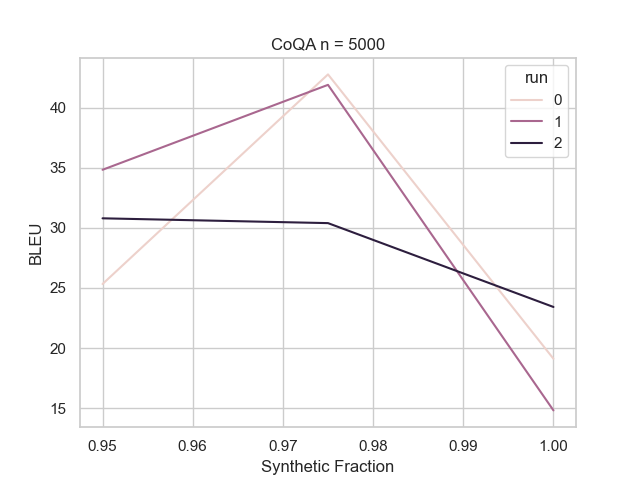}
     \end{subfigure}
     \hfill
     \begin{subfigure}[b]{0.45\textwidth}
         \centering
         \includegraphics[width=\textwidth]{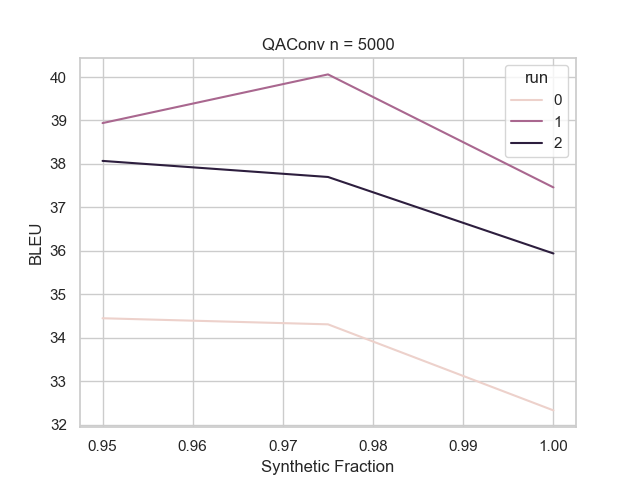}
     \end{subfigure}
        \caption{Model performance as the synthetic proportion of the training data varies from 95\% to 100\%. Across all datasets and random seeds, having \textbf{just 2.5\%} of the training dataset being human generated boosts performance.}
        \label{fig:z5}
\end{figure*}

\noindent\textbf{Main Experiments:} For figures in the main text where only one task is shown (Figure~\ref{fig:full} and Figure~\ref{fig:zoom5joined}), we provide the complete figures with both tasks (Figure~\ref{fig:fulldataboth} and Figure~\ref{fig:zoom5both}). We also provide the individual performance curves for these experiments (Figure~\ref{fig:fullind} and Figure~\ref{fig:z5}).

\noindent\textbf{Robustness to choice of QA metric:}
To verify the robustness of the results, we show that the QA results are not an artifact of the choice of metric (Figure~\ref{table:qafull} and Figure~\ref{table:qaz5}) by using Exact Match, String Inclusion, ROUGE-1~\citep{lin-2004-rouge} and BERTScore~\citep{zhang2019bertscore}. There is overwhelming agreement between all metrics on the rankings of models.  

\noindent\textbf{Addressing spurious correlations:}
We show that the performance gains afforded by human generated data cannot be explained by a spurious correlation between the human generated train and test splits. This would occur when there are significant annotation artifacts that are not relevant to the task, but are correlated with the correct output. We conduct an out-of-domain experiment (Table~\ref{table:distshift}), using different datasets to source the training data and testing on a single hold out dataset. Using more synthetic data leads to performance declines even in the OOD setting, showing that human data is of higher quality and the results from the main text cannot be explained by a spurious correlation between the human test and human training samples. Interestingly, in the OOD setting the decline is steady, and we do not observe the phenomenon of a small amount of human data having a disproportionate impact on  performance. This suggests that the disproportionate impact of human data occurs when the human data is in-domain. We leave a further exploration of the OOD generalization abilities of synthetic vs. human data to future work.

\begin{figure*}[th]   
\centering
\includegraphics[height=0.4\textwidth, width=0.5\textwidth]{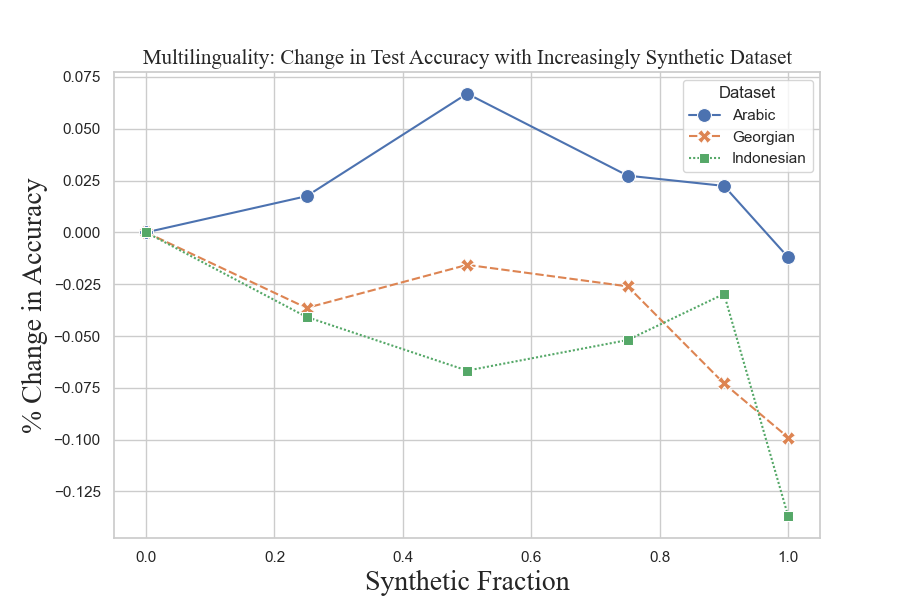}
        \caption{Change in model performance as the proportion of synthetic points in the training data is increased on multilingual fact verification datasets (splits of X-Fact). We observe the same trend as those from earlier experiments, confirming that our results are not limited to the English language. While the phenomenon is reproduced, the threshold of replacement at which we observe a precipitous decline is not the same across languages.}
        \label{fig:multilingual}
\end{figure*}

\noindent\textbf{Multilingual Experiments:}
We replicate our experiment using the Arabic, Georgian, and Indonesian splits of the XFact dataset. We observe (Figure~\ref{fig:multilingual}) the same trend as those from earlier experiments, confirming that our results are not limited to the English language. While the phenomenon is reproduced, the threshold of replacement at which we observe a precipitous decline is not the same across languages. We hypothesize that the language-specific threshold at which a little human data leads to significant performance increases is dependent on how low resource the language is. The study of synthetic data in the multilingual setting has unique considerations that we have not addressed in this work; we leave a focus on these problems to future work. 

\begin{figure*}[t]
     \begin{subfigure}[b]{0.5\textwidth}
         \centering
         \includegraphics[width=\textwidth]{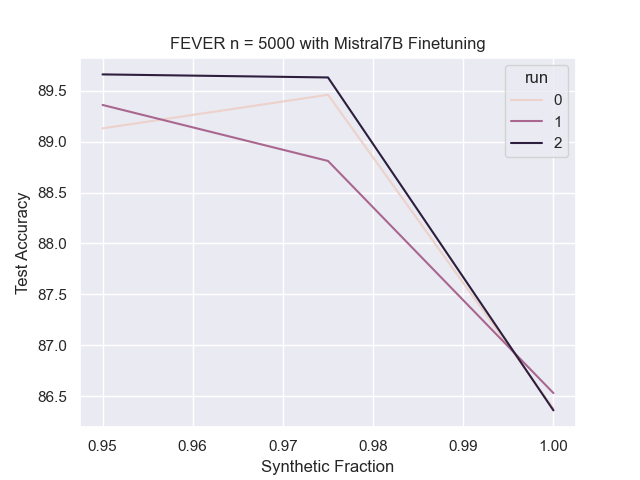}
     \end{subfigure}
     \hfill
     \begin{subfigure}[b]{0.5\textwidth}
         \centering
         \includegraphics[width=\textwidth]{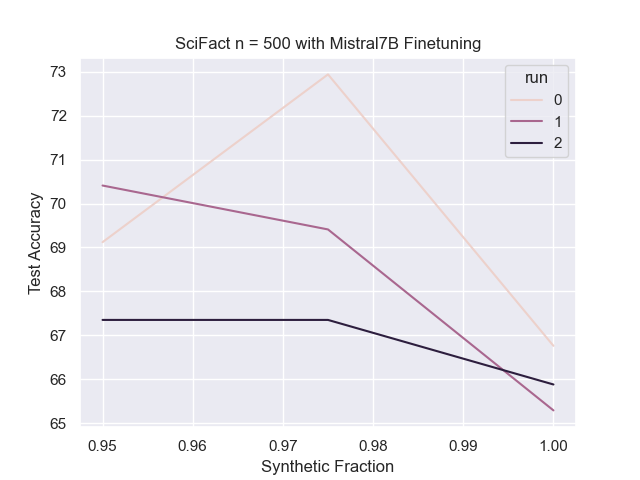}
     \end{subfigure}
     \begin{subfigure}[b]{0.5\textwidth}
         \centering
         \includegraphics[width=\textwidth]{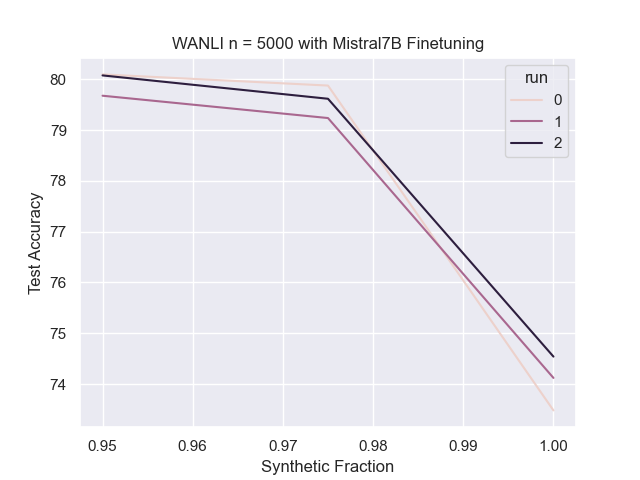}
     \end{subfigure}
          \begin{subfigure}[b]{0.5\textwidth}
         \centering
         \includegraphics[width=\textwidth]{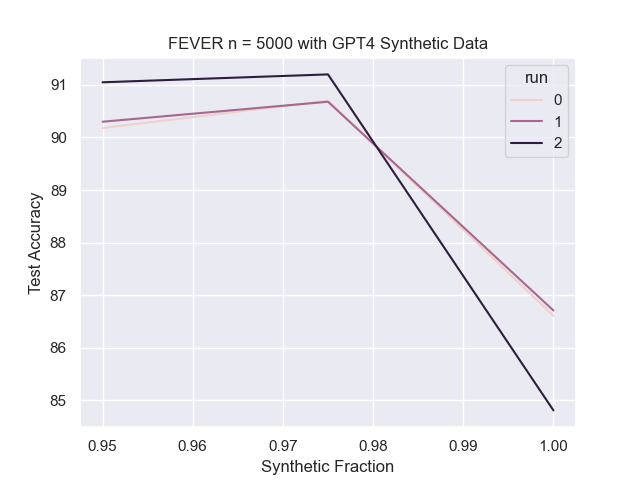}
     \end{subfigure}
     \hfill
     \begin{subfigure}[b]{0.5\textwidth}
         \centering
         \includegraphics[width=\textwidth]{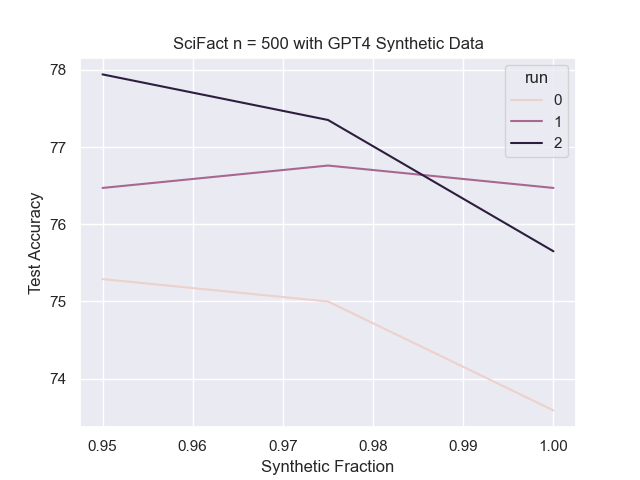}
     \end{subfigure}
     \begin{subfigure}[b]{0.5\textwidth}
         \centering
         \includegraphics[width=\textwidth]{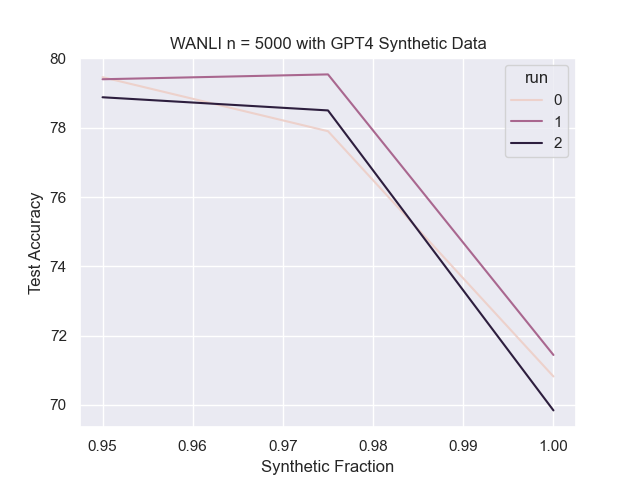}
     \end{subfigure}
        \caption{Results hold consistently on Fact Verification datasets when using Mistral7B as the fine-tuning model and GPT-4 as the prompting model.}
        \label{fig:ablation_fv}
\end{figure*}

\begin{figure*}[t]  
\centering
\includegraphics[height=0.4\textwidth, width=0.5\textwidth]{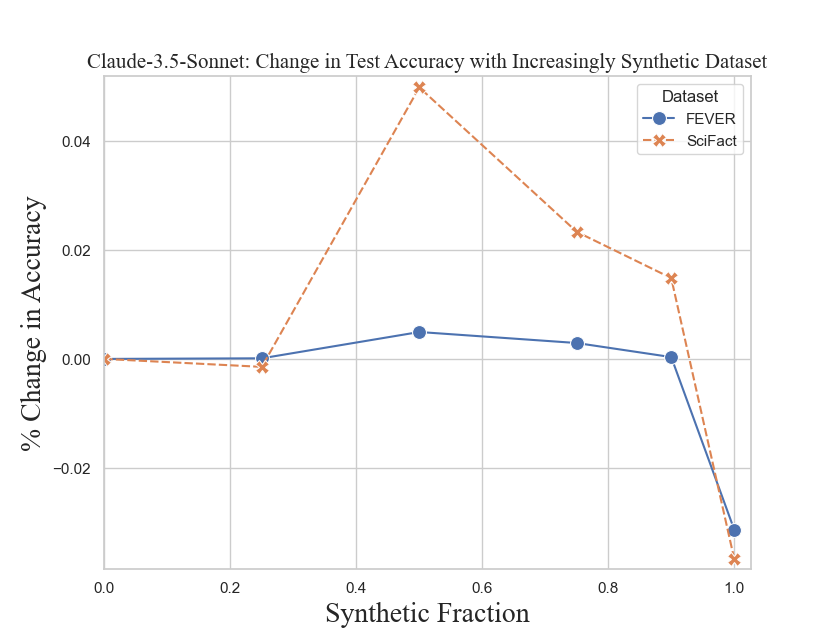}
        \caption{Results hold when using Claude-3.5-Sonnet as the prompting model, showing that the phenomenon is not particular to Synthetic Data from GPT based models.}
        \label{fig:claude}
\end{figure*}

\begin{figure*}[ht]
     \centering
     \begin{subfigure}[b]{0.5\textwidth}
         \centering
         \includegraphics[width=\textwidth]{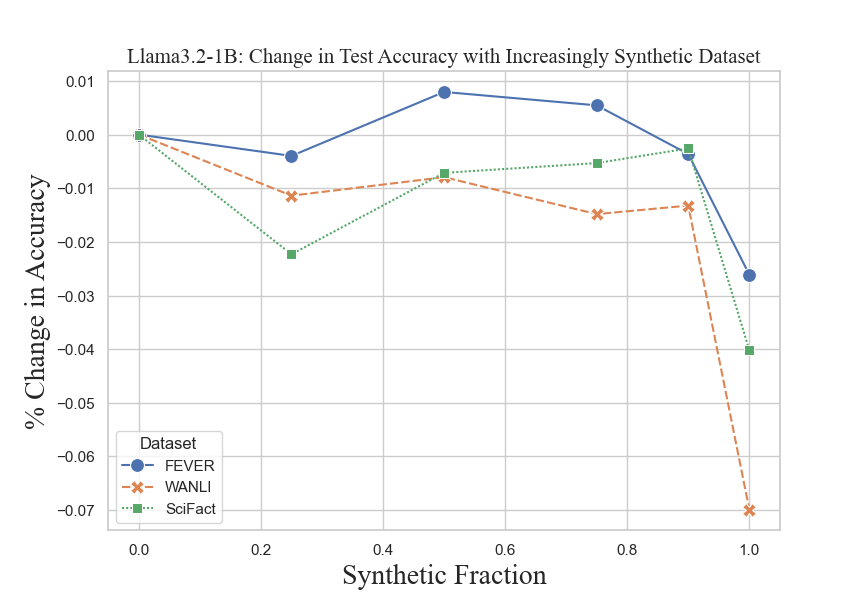}
     \end{subfigure}
     \hfill
     \begin{subfigure}[b]{0.5\textwidth}
         \centering
         \includegraphics[width=\textwidth]{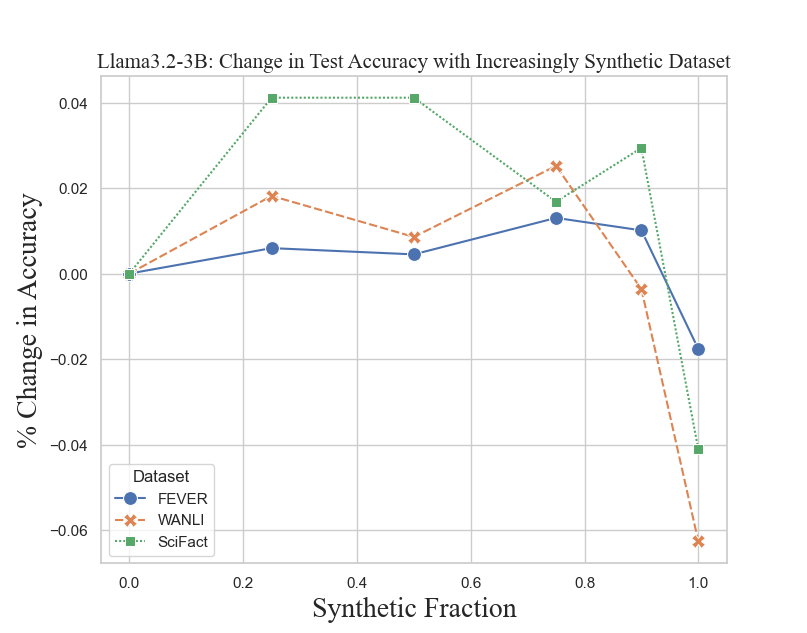}
     \end{subfigure}
     \begin{subfigure}[b]{0.5\textwidth}
         \centering
         \includegraphics[width=\textwidth]{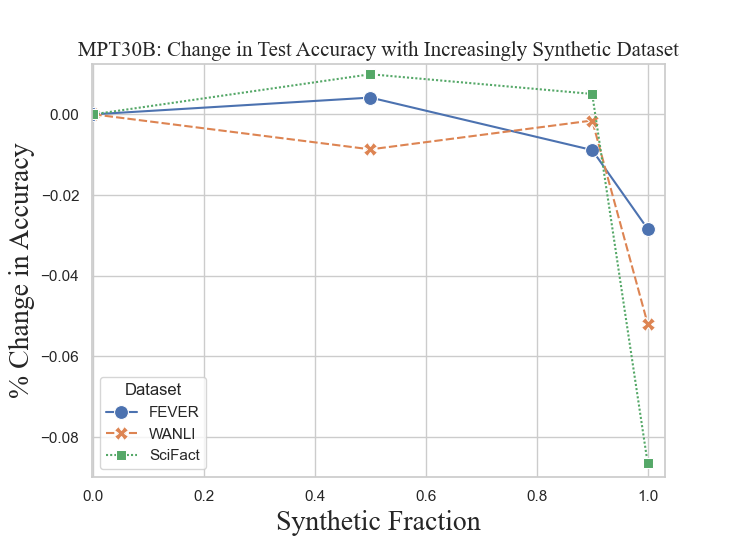}
     \end{subfigure}
        \caption{Results hold consistently on Fact Verification datasets when using models of different scales.}
        \label{fig:scale_ablation}
\end{figure*}

\begin{figure*}[t]
     \begin{subfigure}[b]{0.5\textwidth}
         \centering
         \includegraphics[width=\textwidth]{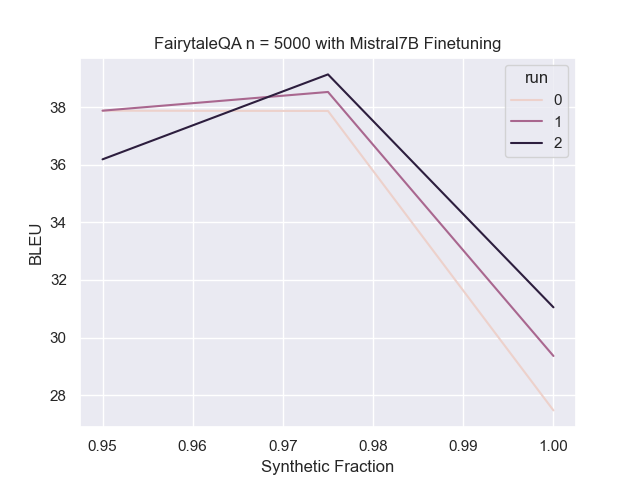}
     \end{subfigure}
     \hfill
     \begin{subfigure}[b]{0.5\textwidth}
         \centering
         \includegraphics[width=\textwidth]{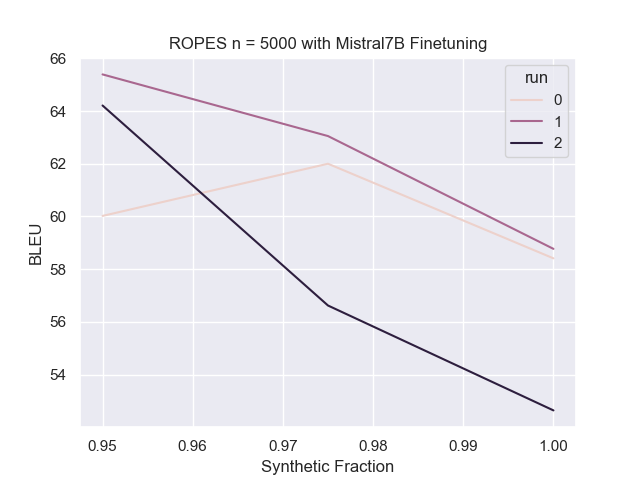}
     \end{subfigure}
     \begin{subfigure}[b]{0.5\textwidth}
         \centering
         \includegraphics[width=\textwidth]{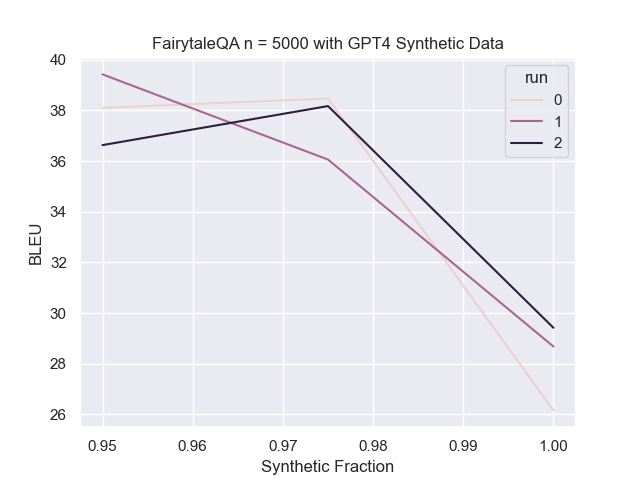}
     \end{subfigure}
          \begin{subfigure}[b]{0.5\textwidth}
         \centering
         \includegraphics[width=\textwidth]{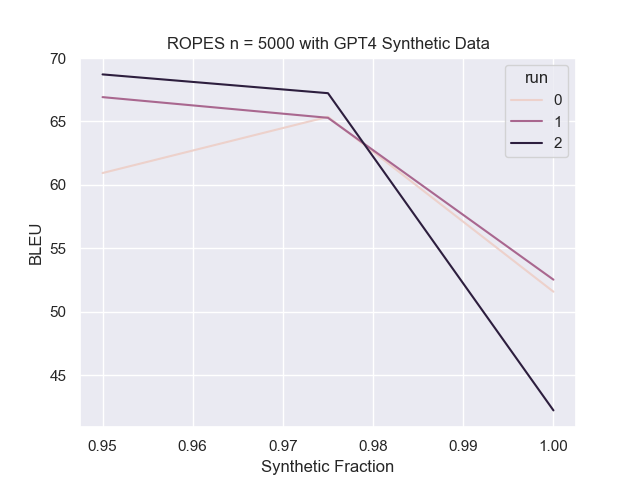}
     \end{subfigure}
        \caption{Results hold consistently on Question Answering datasets when using Mistral7B as the fine-tuning model and GPT-4 as the prompting model}
        \label{fig:ablation_qa}
\end{figure*}

\begin{figure*}[t]
     \centering
     \begin{subfigure}[b]{0.45\textwidth}
         \centering
         \includegraphics[width=\textwidth]{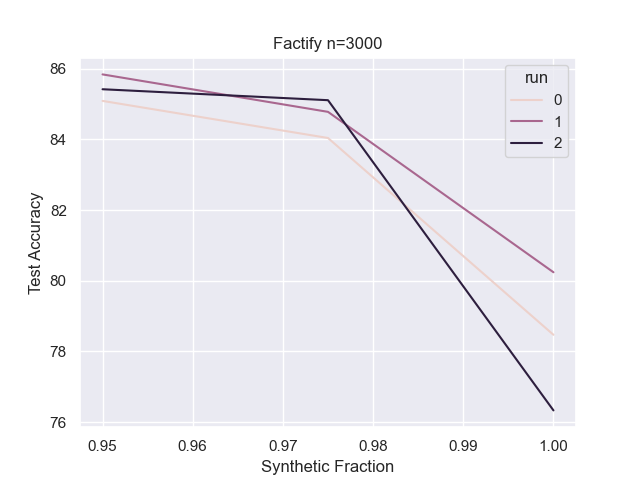}
     \end{subfigure}
     \hfill
     \begin{subfigure}[b]{0.45\textwidth}
         \centering
         \includegraphics[width=\textwidth]{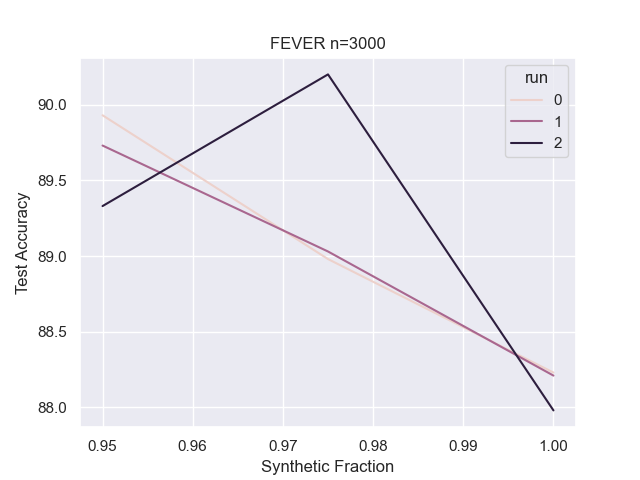}
     \end{subfigure}
     \hfill
     \begin{subfigure}[b]{0.45\textwidth}
         \centering
         \includegraphics[width=\textwidth]{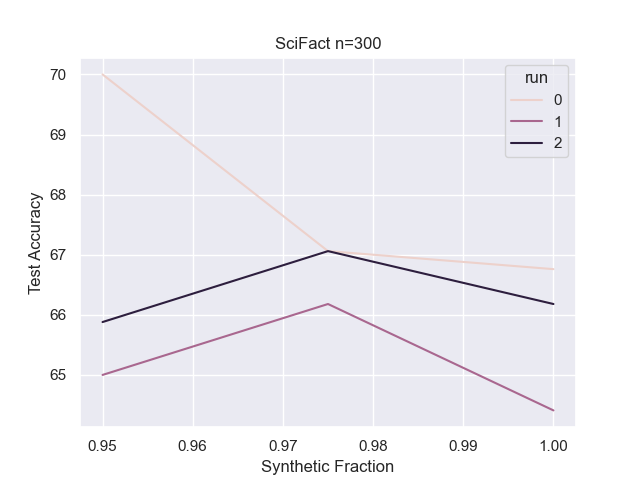}
     \end{subfigure}
     \hfill
     \begin{subfigure}[b]{0.45\textwidth}
         \centering
         \includegraphics[width=\textwidth]{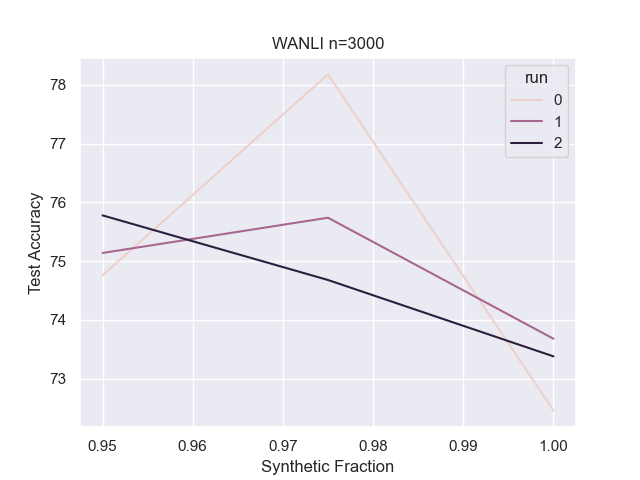}
     \end{subfigure}
      \vfill
     \begin{subfigure}[b]{0.45\textwidth}
         \centering
         \includegraphics[width=\textwidth]{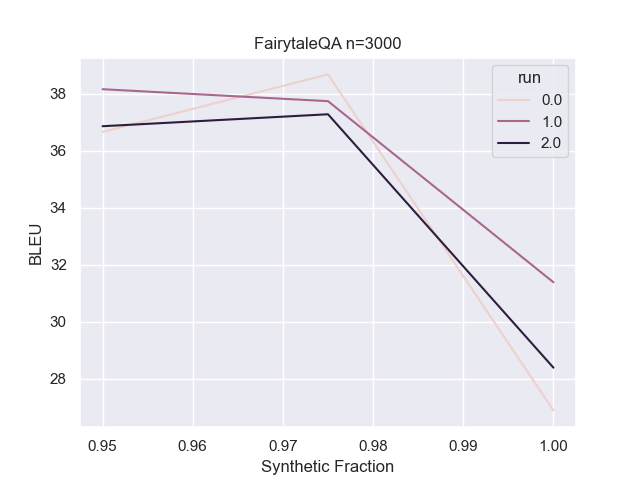}
     \end{subfigure}
     \hfill
     \begin{subfigure}[b]{0.45\textwidth}
         \centering
         \includegraphics[width=\textwidth]{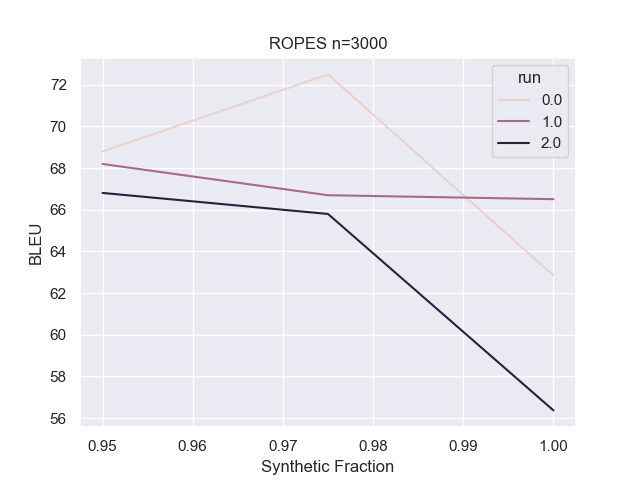}
     \end{subfigure}
        \caption{Model performance as the synthetic proportion of the training data varies from 95\% to 100\% with total number of points $n=3000$. Across all runs on all datasets including \textbf{just 75 real datapoints} can boost performance.}
        \label{fig:z3}
\end{figure*}

\begin{figure*}[t]
     \centering
     \begin{subfigure}[b]{0.45\textwidth}
         \centering
         \includegraphics[width=\textwidth]{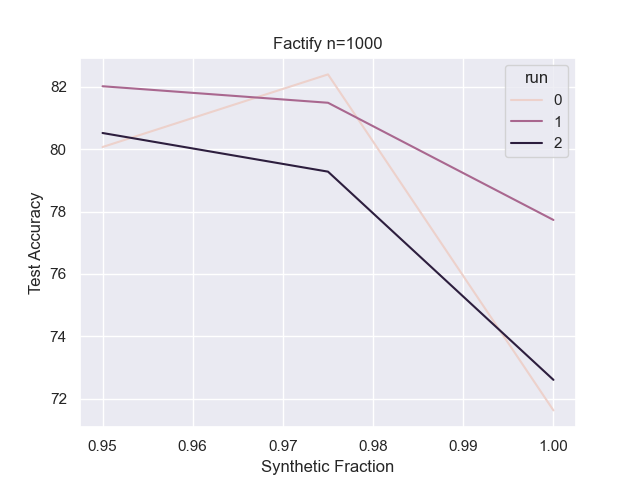}
     \end{subfigure}
     \hfill
     \begin{subfigure}[b]{0.45\textwidth}
         \centering
         \includegraphics[width=\textwidth]{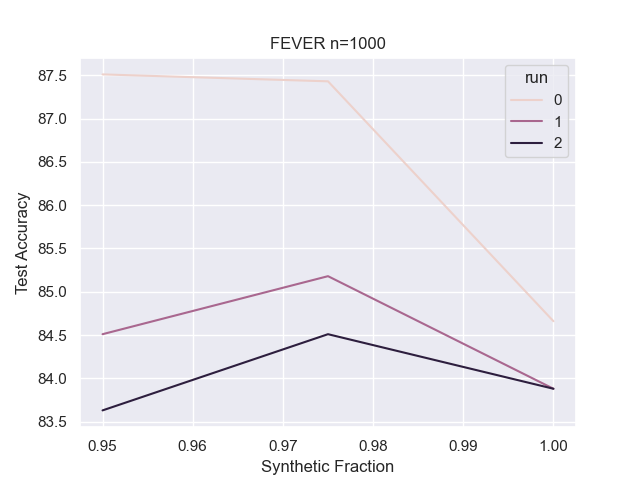}
     \end{subfigure}
     \hfill
     \begin{subfigure}[b]{0.45\textwidth}
         \centering
         \includegraphics[width=\textwidth]{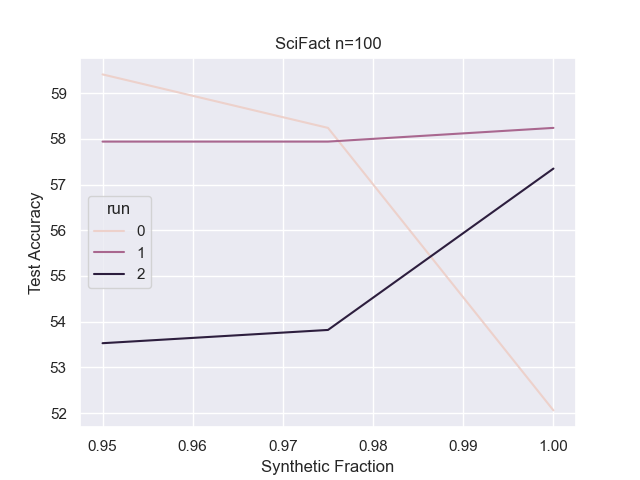}
     \end{subfigure}
     \hfill
     \begin{subfigure}[b]{0.45\textwidth}
         \centering
         \includegraphics[width=\textwidth]{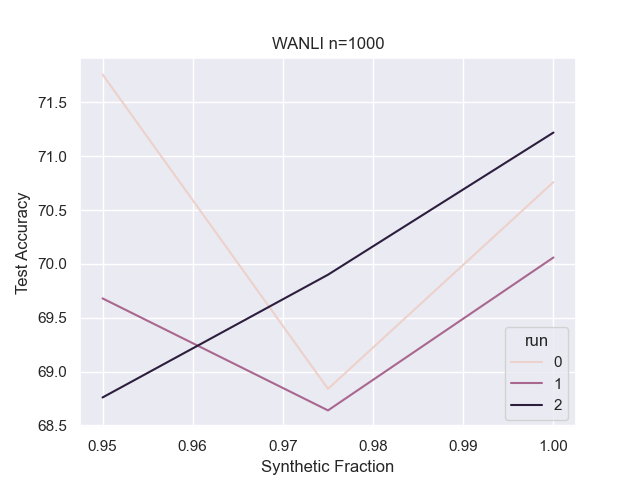}
     \end{subfigure}
      \vfill
     \begin{subfigure}[b]{0.45\textwidth}
         \centering
         \includegraphics[width=\textwidth]{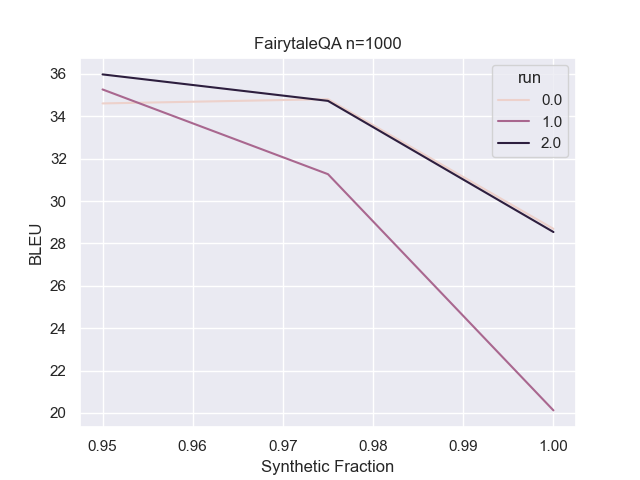}
     \end{subfigure}
     \hfill
     \begin{subfigure}[b]{0.45\textwidth}
         \centering
         \includegraphics[width=\textwidth]{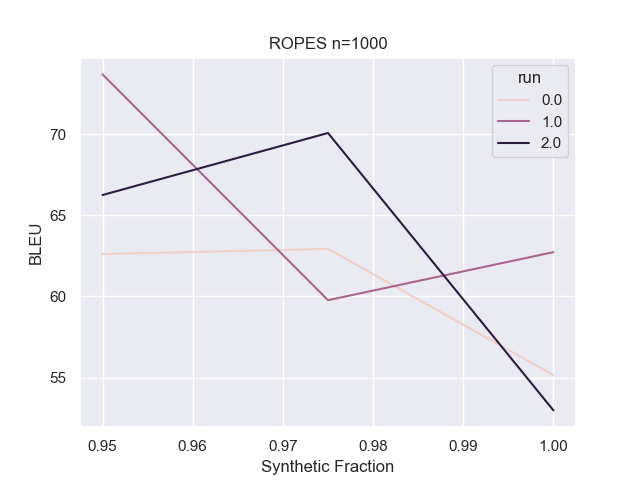}
     \end{subfigure}
        \caption{Model performance as the synthetic proportion of the training data varies from 95\% to 100\% with total number of points $n=1000$. While the most common trend is that including real data improves performance, the results are much more unstable.}
        \label{fig:z1}
\end{figure*}

\noindent\textbf{Ablations:}
We show that the same trends can be seen (\Cref{fig:ablation_fv,fig:claude,fig:scale_ablation,fig:ablation_qa,fig:cot}) when using a different fine-tuning model (Mistral-7B), models of varying scales (from 1B parameter models to 30B parameter models. different prompting models (GPT-4 and Claude-3.5) and a more sophisticated prompting strategy (Chain-of-Thought Prompting). Across all configurations, we see a consistent decrease in performance when moving from 95\% to 100\% synthetic data, confirming that models trained on purely synthetic data can be improved by including just \textbf{125} real data points. For Chain-of-Thought Prompting, the authors manually annotated 3 examples with rationales per dataset to serve as the prompts. The complete examples and pipeline are provided with the code: \url{github.com/dhananjayashok/littlehumandata}

We additionally show that these trends hold across data scales (\Cref{fig:z3,fig:z1}), replicating the experiment with n=3000 and n=1000. While the trend is clearly visible in both cases, the results for n=1000 have more variance and hence have a minority of cases where the relationship does not hold.

\begin{figure*}[ht]
     \begin{subfigure}[b]{0.5\textwidth}
         \includegraphics[width=\textwidth]{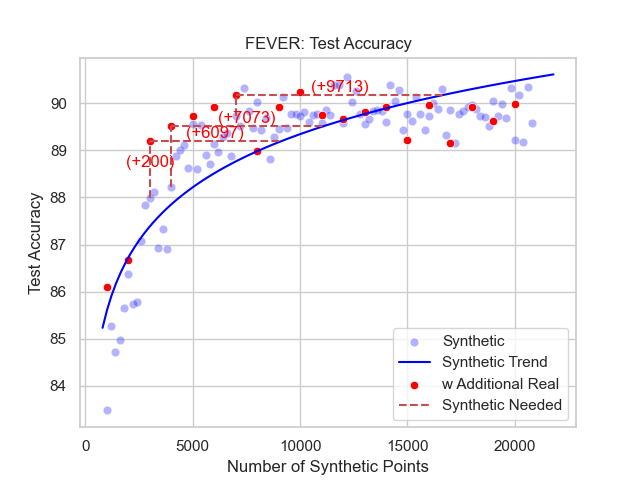}
     \end{subfigure}
     \hfill
     \begin{subfigure}[b]{0.5\textwidth}
         \includegraphics[width=\textwidth]{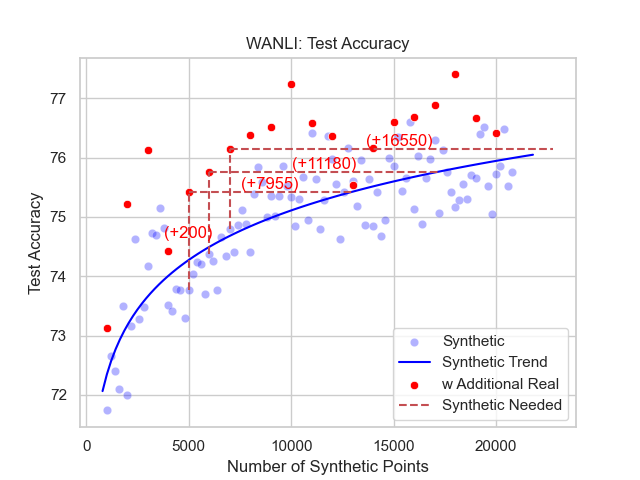}
     \end{subfigure}
     \hfill
     \begin{subfigure}[b]{0.5\textwidth}
         \centering
         \includegraphics[width=\textwidth]{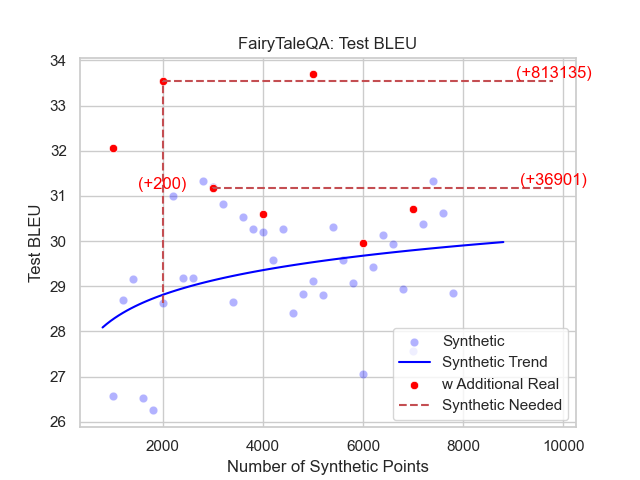}
     \end{subfigure}
     \hfill 
    \begin{subfigure}[b]{0.5\textwidth}
         \centering
         \includegraphics[width=\textwidth]{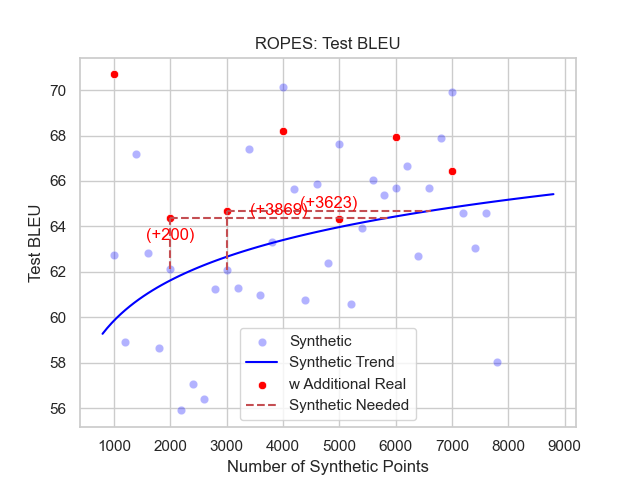}
     \end{subfigure}
        \caption{Adding 200 real data points is as effective as adding an order of magnitude more synthetic data points.}
        \label{fig:money_all}
\end{figure*}

\noindent\textbf{Tradeoff Experiment:} The main text shows results for the experiment detailed in Section~\ref{sec:exp3} on the WANLI dataset (Figure~\ref{fig:money_main}), here we show results on the remaining three datasets (Figure~\ref{fig:money_all}) and provide (Figure~\ref{table:money}) the number of additional synthetic points needed to match the performance gains of 200 additional real points (average, median and standard deviation for each dataset). ROPES shows similar results to WANLI, however FairyTaleQA and FEVER present different trends. On FEVER, we are able to reach the saturation point, after which additional data (whether synthetic or real) does not increase performance. Even in this case, we are able to reach this point of diminishing marginal return more rapidly when using a small amount of synthetic data. On a base synthetic training set of size $3000$, adding $200$ real data points drives the test accuracy to $89.25\%$, a score that is only matched once we add at least 2000 synthetic data points (an order of magnitude larger). On FairyTaleQA, we get enormous estimates for the number of additional synthetic points needed (a mean of 2.8e5). We do not interpret these numbers literally, rather seeing this as a sign that human generated data may occasionally boost performance to an extent that could be fundamentally unachievable by purely synthetic data.

\section{Synthetic Data Generation}
\label{sec:appendix_synthetic}
\begin{figure*}
    \centering
    \includegraphics[width=0.7\linewidth]{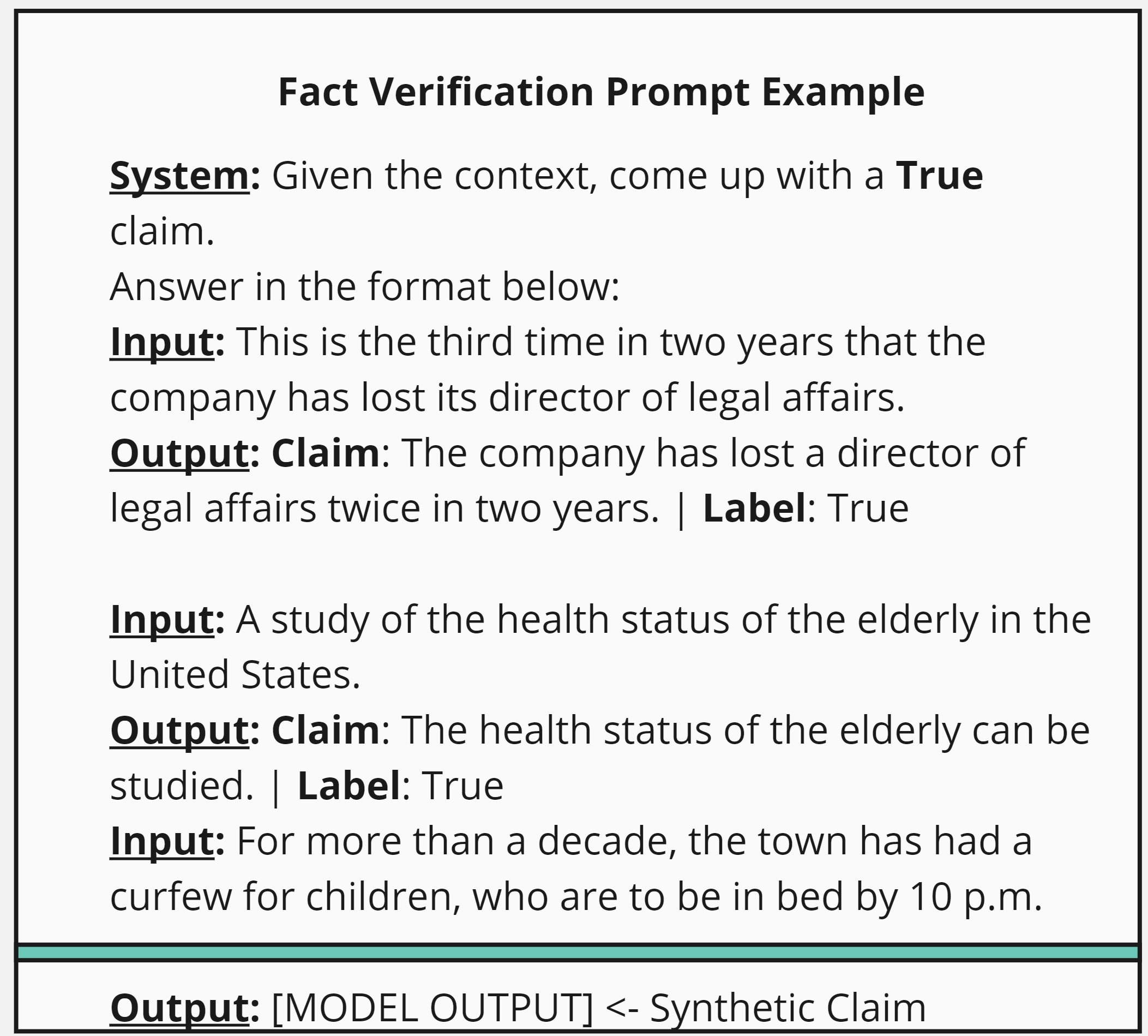}
    
    \vspace{0.02\linewidth}
    \includegraphics[width=0.7\linewidth]{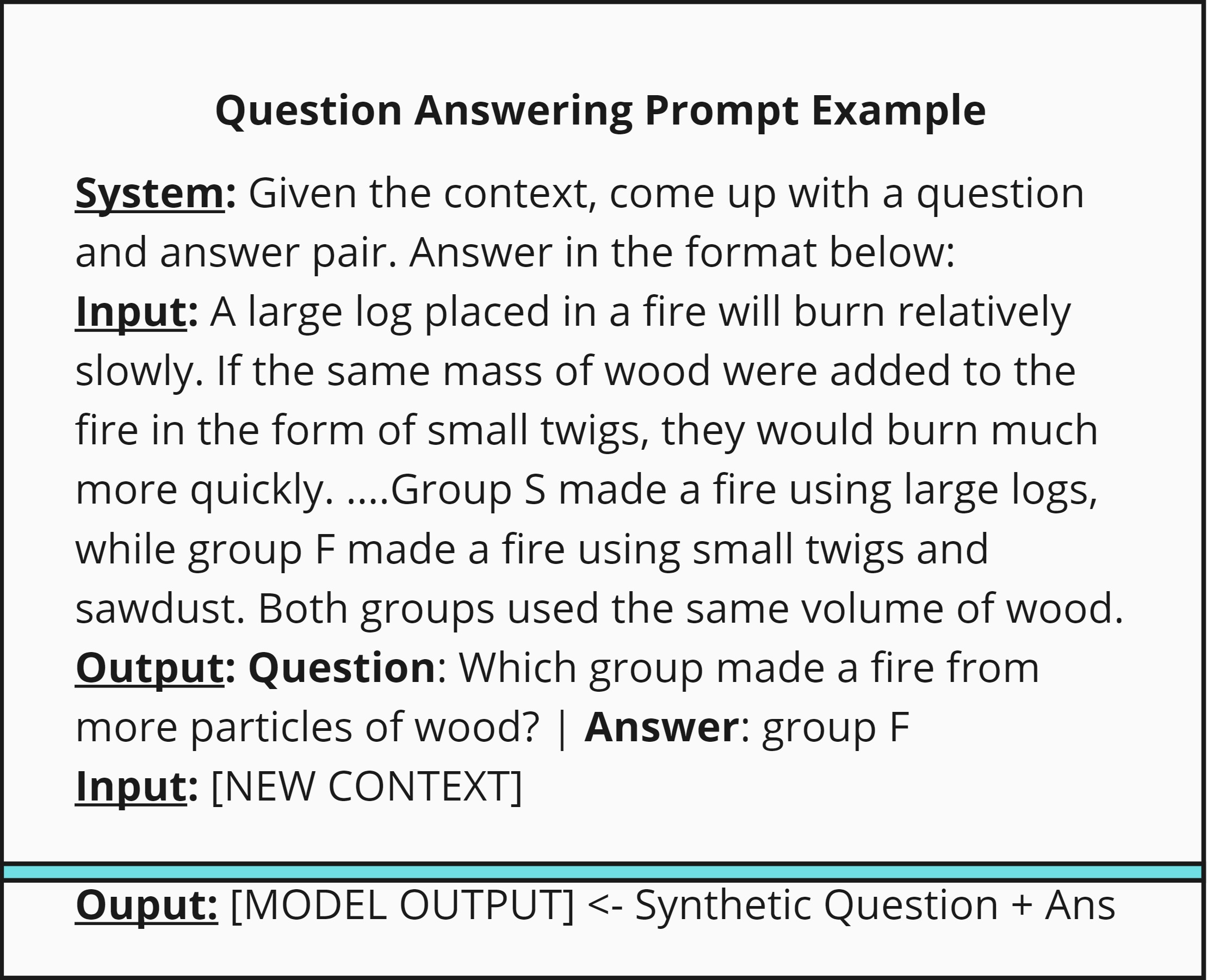}
    \caption{Example prompts used to synthetically generate (claim, label) or (question, answer) pairs using a new context / evidence text.}
    \label{fig:prompts}
\end{figure*}

In our implementation (Figure~\ref{fig:prompts}), we use few-shot learning with $k=3$, i.e., three  examples per query, with each example drawn randomly (with replacement) from the training set of the specific dataset. 

We generate one synthetic point for every real point in the dataset, using the evidence text it is associated with. This gives us a total of $n$ synthetic data points for every $n$ real data points in a dataset. 

We observed that if we did not correct for label shift, the prompt model would be heavily biased towards True claims, i.e., it would generate a dataset containing 90\% True claims, while original datasets have proportions between 33\%--60\% True. 

For the synthetic datasets used in our experiments, we correct for this label shift by specifying the label of the claim we wish to generate and providing only examples of claims with that specific label in the prompt. 

For all datasets, we verify that the diversity of the generated claims/questions/answers are comparable to that of the human generated texts (see Appendix~\ref{sec:appendix_dicussion})

\begin{figure*}[t]
     \centering
     \begin{subfigure}[b]{0.45\textwidth}
         \centering
         \includegraphics[width=\textwidth]{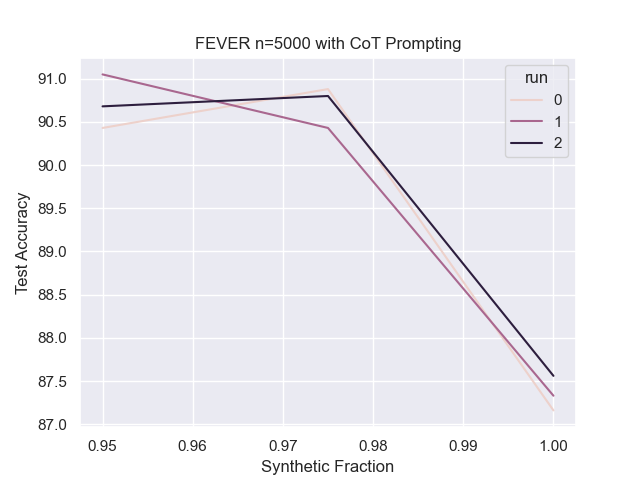}
     \end{subfigure}
     \hfill
     \begin{subfigure}[b]{0.45\textwidth}
         \centering
         \includegraphics[width=\textwidth]{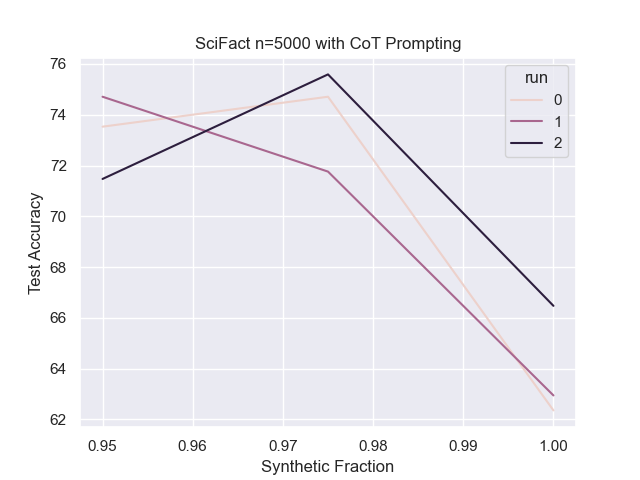}
     \end{subfigure}
     \hfill
     \begin{subfigure}[b]{0.45\textwidth}
         \centering
         \includegraphics[width=\textwidth]{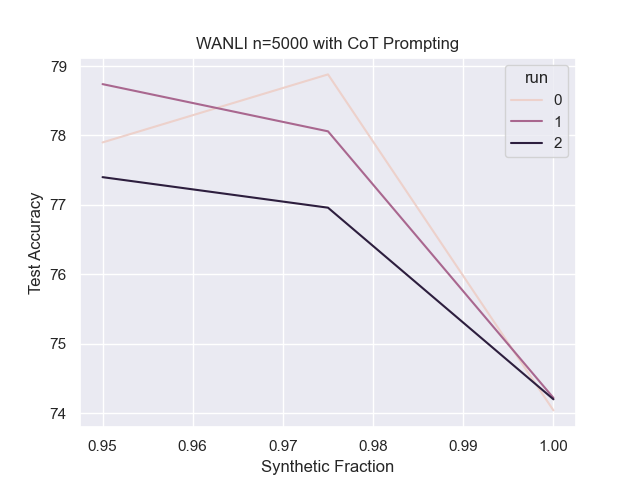}
     \end{subfigure}
      \vfill
     \begin{subfigure}[b]{0.45\textwidth}
         \centering
         \includegraphics[width=\textwidth]{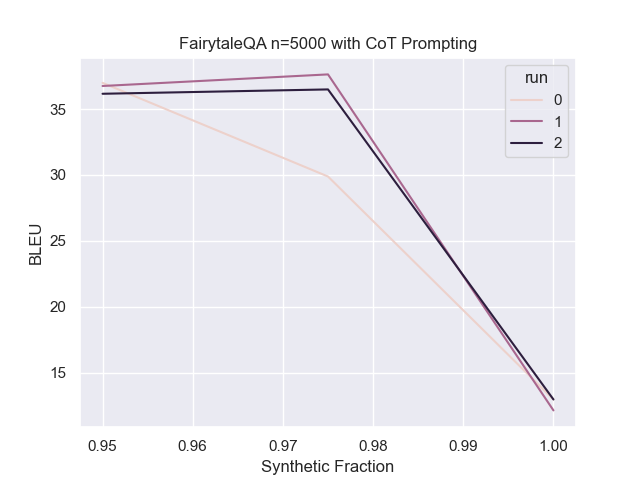}
     \end{subfigure}
     \hfill
     \begin{subfigure}[b]{0.45\textwidth}
         \centering
         \includegraphics[width=\textwidth]{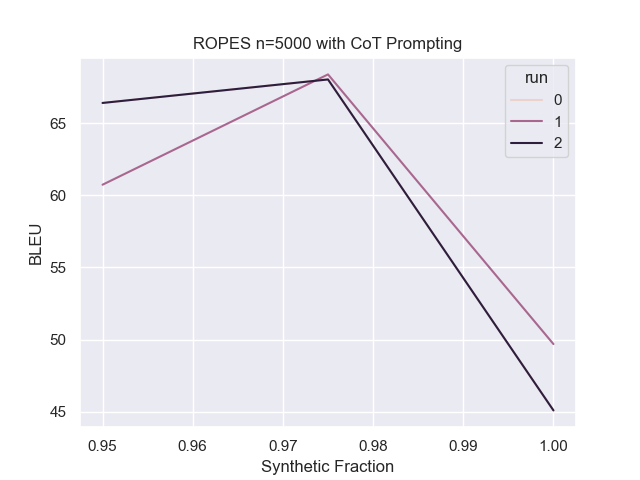}
     \end{subfigure}
        \caption{Results hold when using Chain-Of-Thought Prompting on GPT-3.5}
        \label{fig:cot}
\end{figure*}

This setting is generous towards synthetic data generation. In practice, we might only have three fixed examples to use in the prompt, potentially reducing the diversity of synthetic data generated. We verify that this does not affect the results of our experiment in the Chain-of-Thought ablation (Figure~\ref{fig:cot}), where we use a fixed set of examples to generate all synthetic data points. We may also not know the correct label proportion to ask for and suffer a significant label shift when using synthetic data generation.

\section{Datasets Used}
\label{sec:appendix_datasets}

All datasets used below are released under open use licenses, authorizing their use in this research. For each dataset, we discuss the potential of dataset leakage (i.e., whether the data has been exposed to GPT-3.5-Turbo during its training) as well as the extent of automation involved in the generation of each dataset. However, across all experiments and ablations, these factors do not seem to have any discernable effect on the trends discovered in this work. 

\subsection{Fact Verification Datasets}

\noindent\textbf{FEVER }\citep{thorne2018fever} is a dataset of claims about specific entities, generated by altering sentences extracted from Wikipedia. The evidence passages are sentences from Wikipedia articles relevant to the entity in question. This dataset has been well established for a long time before the release of the prompting models used in this work, increasing the chance that it has been exposed to the prompt model ahead of time.

\noindent\textbf{SciFact} \citep{wadden2020fact} is a fact verification dataset for the scientific domain, which uses the abstracts of scientific articles as evidence texts. The corpus is collected from S2ORC~\citep{lo2019s2orc}, a publicly-available corpus of millions of
scientific articles. Annotators are shown a source citation
in the context of an article, and are asked to write up
to three claims based on the content of the citation. 

The above datasets are popular NLP challenge sets that were well known even before the release of GPT-3.5-Turbo~\citep{brown2020language}, the prompting model used in this work. The following two datasets were released after the official training date cut-off, guaranteeing that the data has not been seen ahead of time.  

\noindent\textbf{WANLI}~\citep{Liu2022WANLIWA} is an NLI dataset of 108K examples created through a hybrid worker and AI collaboration approach. The creators first study MultiNLI~\citep{MNLI} and use dataset cartography to automatically identify examples that demonstrate challenging reasoning patterns. 
and then instruct GPT-3 to compose new examples with similar patterns. Machine generated
examples are then automatically filtered, and
finally revised and labeled by human crowd workers. While GPT-3.5-Turbo has not been trained on this data, it is worth noting that the data is partially synthetically generated. 

\noindent\textbf{FACTIFY}~\citep{Mishra2022FACTIFYAM} is a dataset on multi-modal fact verification. It contains images, textual claims, reference textual documents and reference images. The dataset marks some examples that can be verified using text only; we use this sample in our experiments. This dataset was released after the training cut-off date for GPT-3.5 and takes its evidence textsclaims from human-written news or editorial articles. This ensures that the prompt models studied have not seen the data before training. 

\noindent\textbf{Label mapping for NLI and FV:}
While all of the above datasets contain labels for Supports, Refutes, and Not Enough Information (or Entails, Contradicts, Neutral), we consider the stricter formulation of Fact Verification used by~\citet{Honovich2022TRUERF} and ~\citet{Zha2023AlignScoreEF}, considering a claim to be factual if the label is Supports (Entails), and non-factual otherwise. 

 \subsection{Question Answering Datasets}

\noindent\textbf{ROPES} \citep{Lin2019ReasoningOP} is a QA dataset which tests a system's ability to apply knowledge from a passage of text to a new situation. The evidence context contains causal or qualitative relation(s) (e.g., ``animal pollinators increase efficiency of fertilization in flowers''), and a novel situation that uses this background. The question requires reasoning about effects of the relationships in the background passage in the context of the situation.

\noindent\textbf{CoQA} \citep{reddy2019coqa}  is a dataset for building Conversational Question Answering systems. CoQA measures the ability of machines to understand a text passage and answer a series of interconnected questions that appear in a conversation. In our experiments, we extract only the first question in the series and use this to obtain our (context, question, answer) data points. 

\noindent\textbf{QAConv} \citep{Wu2021QAConvQA} focuses on informative conversations, including business emails, panel discussions, and work channels. The creators collect QA pairs with both human-written and machine-generated questions. They use a question generator and a dialogue summarizer as auxiliary tools to collect and recommend questions. While the arXiv version of the paper appears before the GPT-3 cut-off data (April 2021 to the cut-off date of Sept 2021), the paper itself appeared only at ACL 2022. It is still possible that the training data was compromised, and owing to the lack of clarity on the training data used for GPT-3 we have no way to confirm or deny this speculation. 

\noindent\textbf{FairyTaleQA} \citep{xu2022fantastic} is a dataset focusing on narrative comprehension of kindergarten to eighth-grade students. The evidence texts are derived from children-friendly stories which serve as evidence texts. The questions are both explicit and implicit, covering seven types of narrative elements or relations. This dataset was released after the GPT-3 training cut-off date, ensuring that it has not been seen by our prompt model before.

\section{Detailed Discussion on Differences Between Synthetic and Human Data}
\label{sec:appendix_dicussion} 
\begin{figure*}[ht]
     \begin{subfigure}[b]{0.45\textwidth}
         \includegraphics[width=\textwidth]{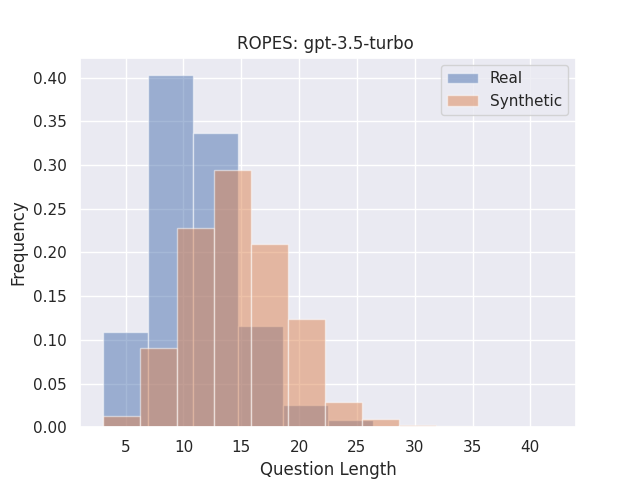}
     \end{subfigure}
     \hfill
     \begin{subfigure}[b]{0.45\textwidth}
         \includegraphics[width=\textwidth]{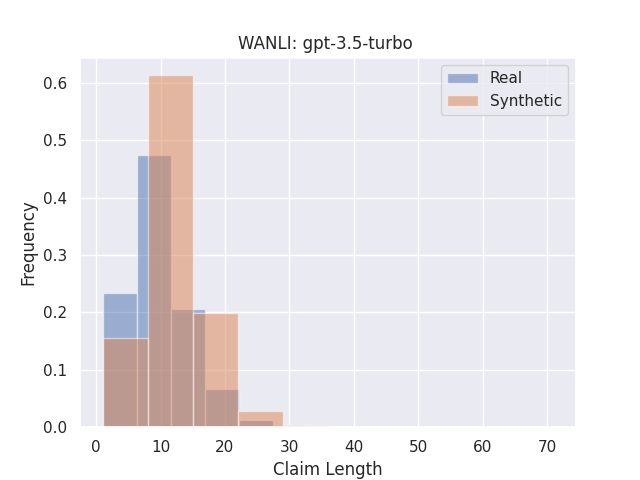}
     \end{subfigure}
     \hfill
     \begin{subfigure}[b]{0.45\textwidth}
         \centering
         \includegraphics[width=\textwidth]{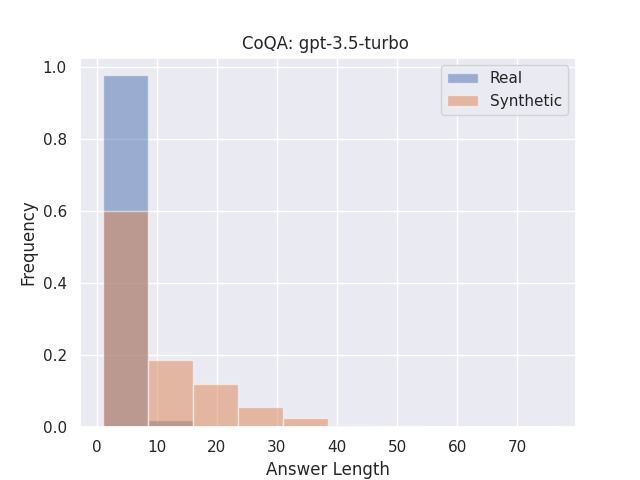}
     \end{subfigure}
     \hfill
     \begin{subfigure}[b]{0.45\textwidth}
         \centering
         \includegraphics[width=\textwidth]{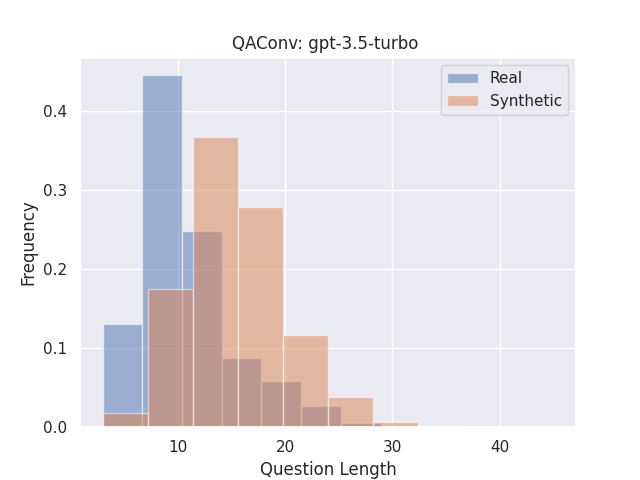}
     \end{subfigure}
      \vfill
     \begin{subfigure}[b]{0.45\textwidth}
         \centering
         \includegraphics[width=\textwidth]{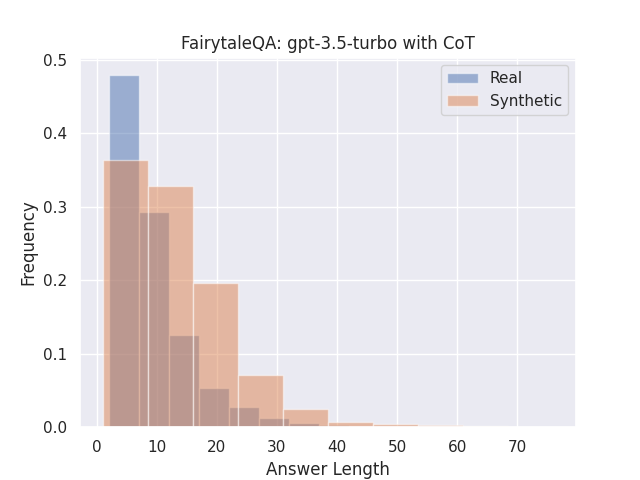}
     \end{subfigure}
     \hfill
     \begin{subfigure}[b]{0.45\textwidth}
         \centering
         \includegraphics[width=\textwidth]{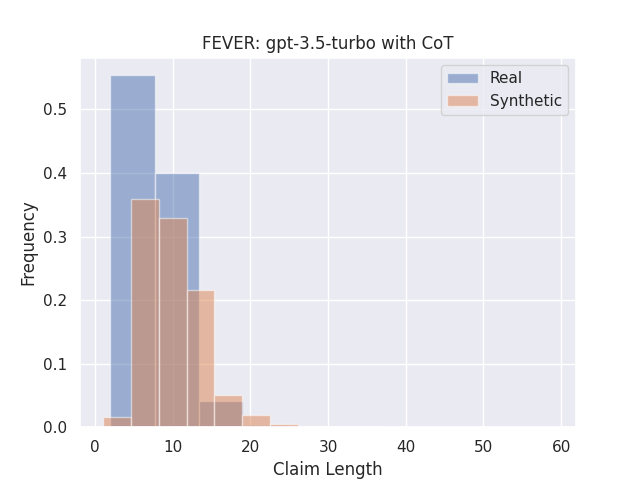}
     \end{subfigure}
     \hfill
     \begin{subfigure}[b]{0.45\textwidth}
         \centering
         \includegraphics[width=\textwidth]{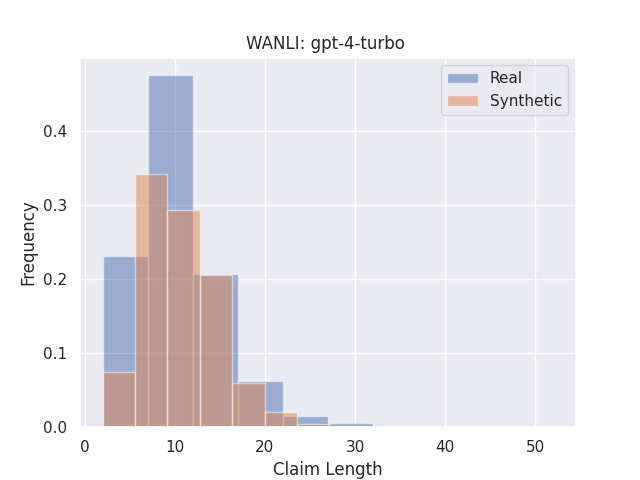}
     \end{subfigure}
     \hfill
     \begin{subfigure}[b]{0.45\textwidth}
         \centering
         \includegraphics[width=\textwidth]{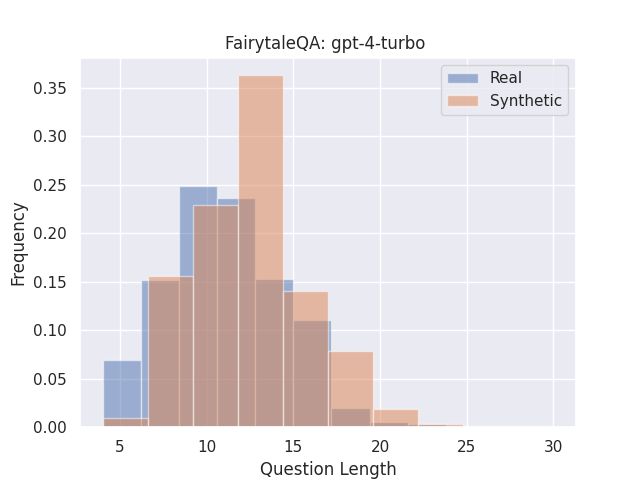}
     \end{subfigure}
        \caption{Synthetic data is, on average, longer than its human generated counterpart. This trend can be seen on FV (claims) and QA (claims and questions), and holds across prompt models and strategies.}
        \label{fig:size_analysis}
\end{figure*}

\begin{figure*}[ht]
     \begin{subfigure}[b]{0.45\textwidth}
         \includegraphics[width=\textwidth]{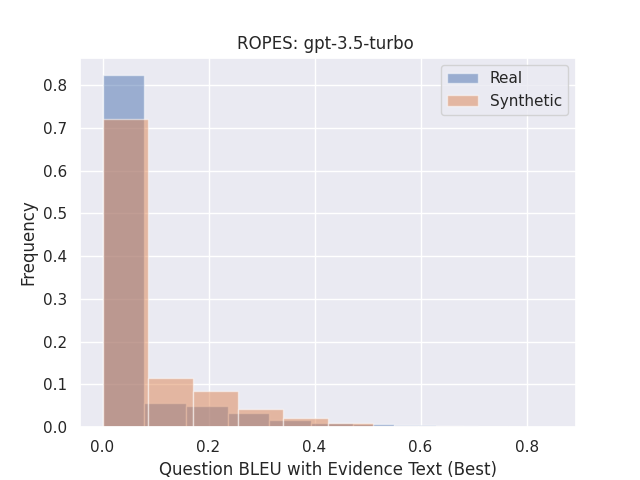}
     \end{subfigure}
     \hfill
     \begin{subfigure}[b]{0.45\textwidth}
         \includegraphics[width=\textwidth]{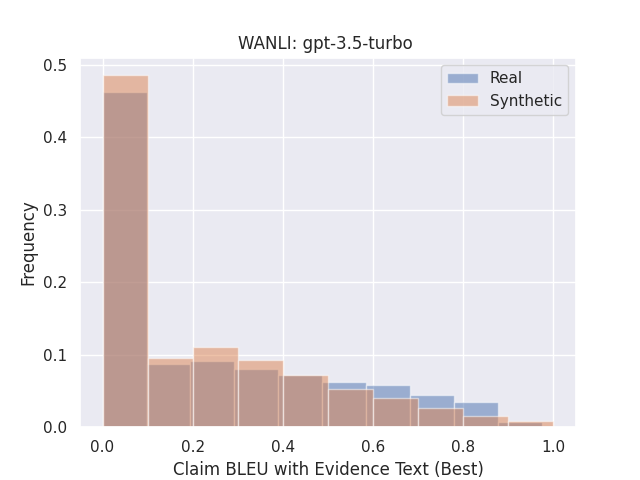}
     \end{subfigure}
     \hfill
     \begin{subfigure}[b]{0.45\textwidth}
         \centering
         \includegraphics[width=\textwidth]{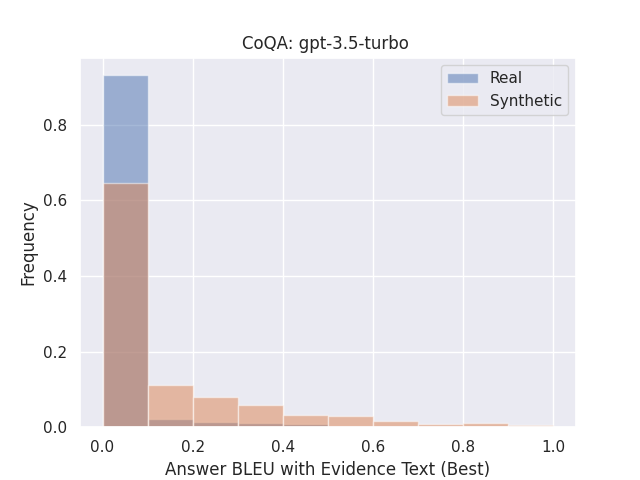}
     \end{subfigure}
     \hfill
     \begin{subfigure}[b]{0.45\textwidth}
         \centering
         \includegraphics[width=\textwidth]{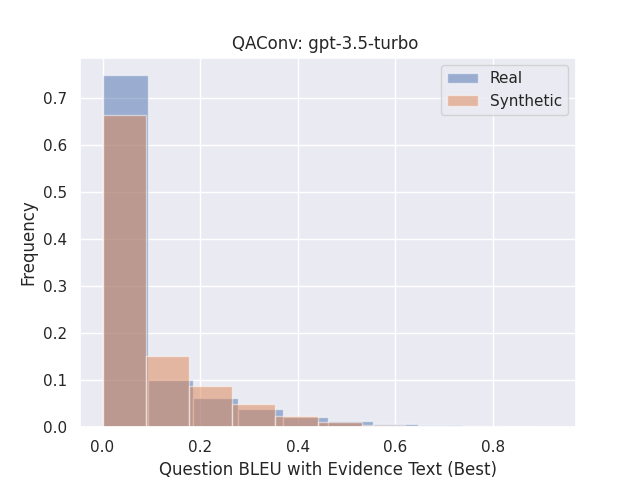}
     \end{subfigure}
      \vfill
     \begin{subfigure}[b]{0.45\textwidth}
         \centering
         \includegraphics[width=\textwidth]{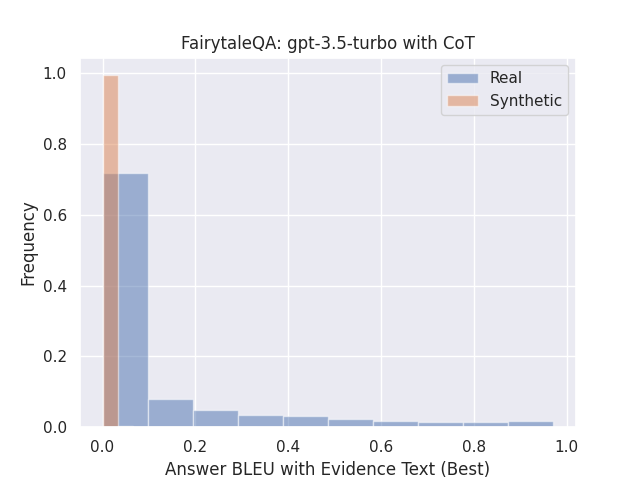}
     \end{subfigure}
     \hfill
     \begin{subfigure}[b]{0.45\textwidth}
         \centering
         \includegraphics[width=\textwidth]{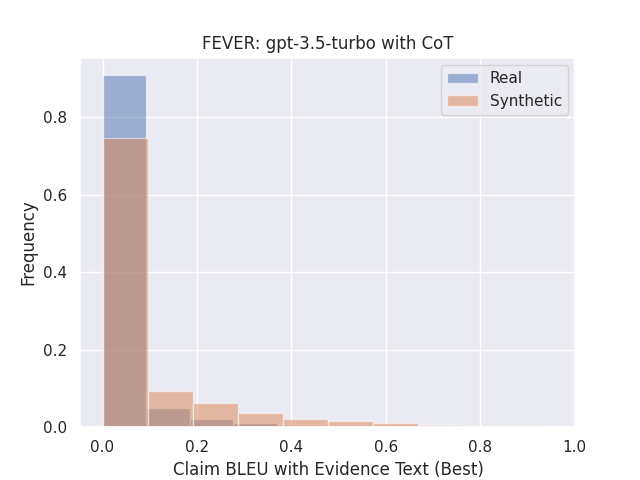}
     \end{subfigure}
     \hfill
     \begin{subfigure}[b]{0.45\textwidth}
         \centering
         \includegraphics[width=\textwidth]{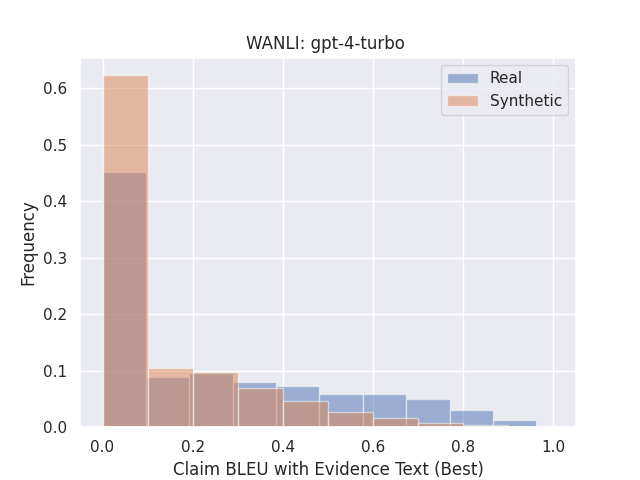}
     \end{subfigure}
     \hfill
     \begin{subfigure}[b]{0.45\textwidth}
         \centering
         \includegraphics[width=\textwidth]{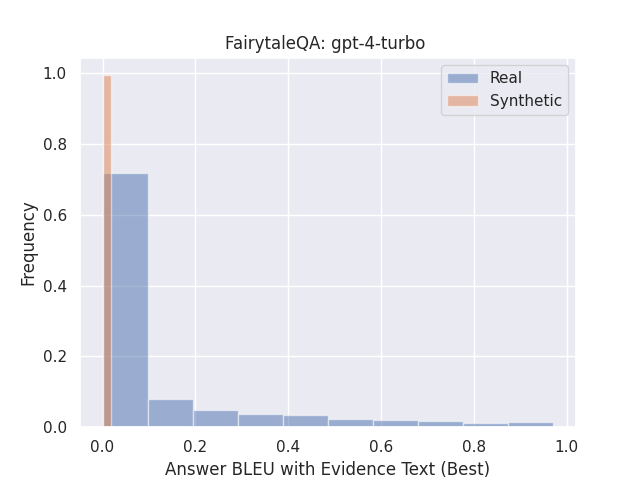}
     \end{subfigure}
        \caption{Synthetic data generally exhibits a higher maximum BLEU score measured against sentences from the context. This suggests that synthetic questions, answers, and claims are more extractive than their human generated counterparts}
        \label{fig:extractive_analysis}
\end{figure*}

\begin{figure*}[ht]
     \begin{subfigure}[b]{0.5\textwidth}
         \includegraphics[width=\textwidth]{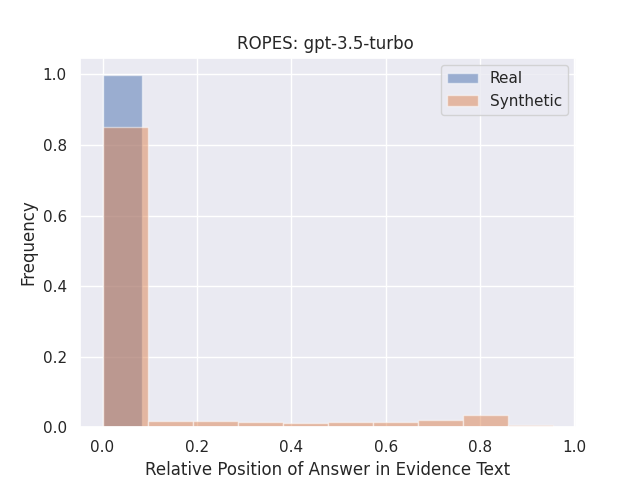}
     \end{subfigure}
     \hfill
     \begin{subfigure}[b]{0.5\textwidth}
         \includegraphics[width=\textwidth]{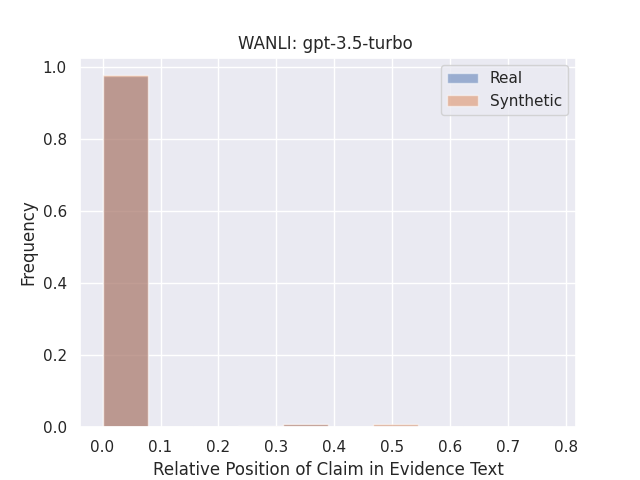}
     \end{subfigure}
     \hfill
     \begin{subfigure}[b]{0.5\textwidth}
         \centering
         \includegraphics[width=\textwidth]{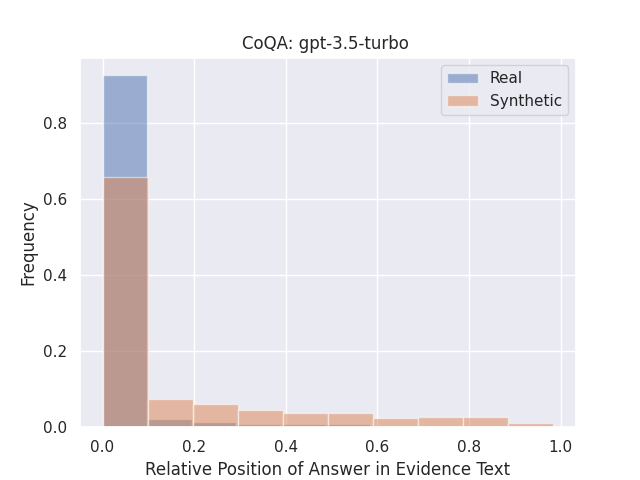}
     \end{subfigure}
     \hfill
     \begin{subfigure}[b]{0.5\textwidth}
         \centering
         \includegraphics[width=\textwidth]{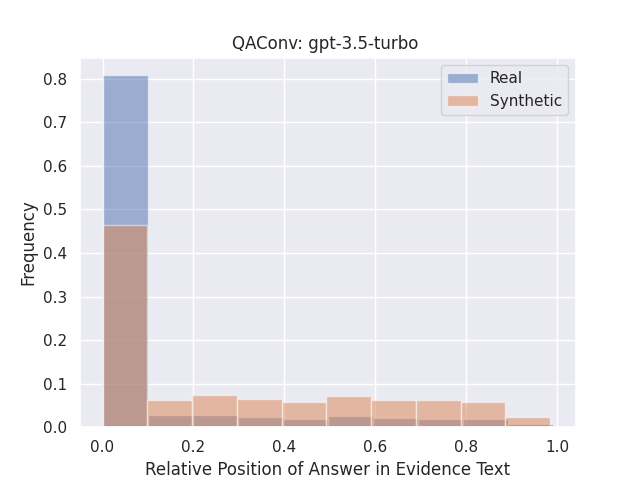}
     \end{subfigure}
      \vfill
     \begin{subfigure}[b]{0.5\textwidth}
         \centering
         \includegraphics[width=\textwidth]{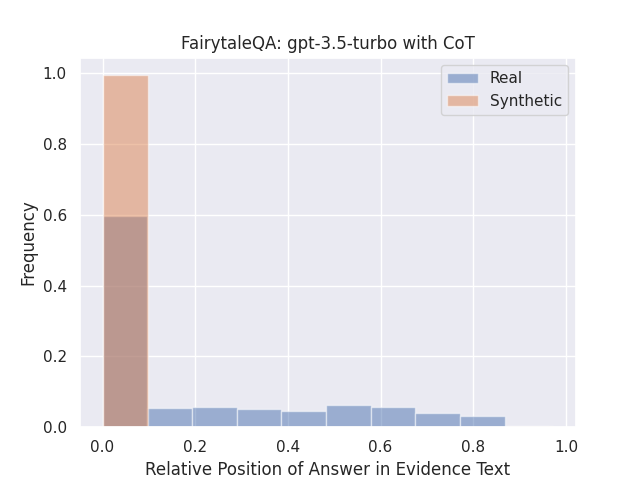}
     \end{subfigure}
     \hfill
     \begin{subfigure}[b]{0.5\textwidth}
         \centering
         \includegraphics[width=\textwidth]{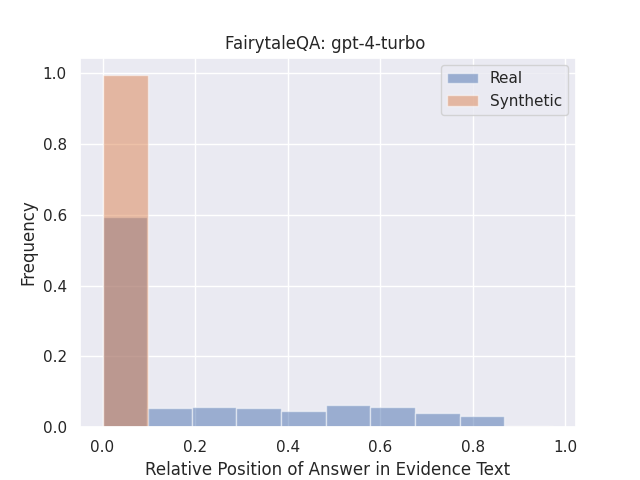}
     \end{subfigure}
        \caption{Synthetic data typically chooses more diverse sources (in terms of answer location or claim location in the evidence text), while humans tend to favor the start of the evidence text.}
        \label{fig:position_analysis}
\end{figure*}

To compute the extent to which the evidence sentences `contain' the questions, answers, and claims, we measure the BLEU of the generation with \textbf{each individual sentence} of the evidence texts, plotting the maximum of these BLEU scores in Figure~\ref{fig:extractive_analysis}.
We find that synthetic generations have a far higher n-gram overlap with the evidence sentences than human generations. This suggests that synthetic data generation produces data points that are more extractive, while humans are more likely to abstract from the evidence. We also use the position of the evidence sentence that achieves the highest BLEU score as a proxy for the source location of the synthetic generation, and find that synthetic data generation chooses more diverse sources for the question and answer content, with human annotation overwhelmingly more likely to create questions whose answers lie in the start of the evidence texts (Figure~\ref{fig:position_analysis}). Finally, the main text shows the size length comparison for a single dataset. Here we provide a larger sample 
(Figure~\ref{fig:size_analysis}). We explore the errors created by the models trained on 0\% and 100\% data, searching for trends or divergences between the input instances that achieve a low prediction accuracy or score. Our investigation finds no major distinguishing factors between them, leaving a more fine-grained study of the effect of purely synthetic data on model decision-making to future work. 

\section{Implementation Details}
\label{sec:appendix_implementation}
While our full code implementation can be seen in the GitHub repository (to be released after review), we list the key implementation details below. 

\textbf{Hardware and Systems Used}: The experiments were run on a cluster that included nodes with: five A40 GPUs (48GB), three RTX 2080Tis, and a separate machine using a single A100 GPU. 

\textbf{Prompt Models used}: We used GPT-3.5-Turbo and GPT-4-Turbo Batch APIs from OpenAI. Generations were obtained at various points from  August 2024 to September 2024. 

\textbf{Fine-Tuning Models Used}: We used two fine-tuning models in our experiments. Llama3 used the Llama3.1-8B HuggingFace Checkpoint, and Mistral used the Mistral7B-Instruct-v0.2 HuggingFace Checkpoint. We did not conduct an extensive hyperparameter search, however we tried various epochs on smaller samples of the FEVER and ROPES datasets, selecting that number for every dataset on all experiments. Fact verification models used Adam Optimization with a learning rate of 1e-5 for two epochs, while QA datasets used  a learning rate of 1e-2 for five epochs.

\begin{table*}[h]
\begin{tabular}
{@{}ll||rrrrrr@{}}
\toprule
\textbf{Dataset} & \textbf{Synthetic \%} & \textbf{EM} & \textbf{Inc} & \textbf{R Inc} & \textbf{BLEU} & \textbf{ROUGE} & \textbf{BERTScore} \\ \midrule
\multirow{6}{*}{CoQA}             & 0                           & 40.6                 & 52.2               & 60.8                      & 47.9         & 64.0            & 87.8              \\
             & 25                        & 35.0                   & 51.8               & 54.8                       & 44.1         & 62.0            & 85.9              \\
             & 50                         & 31.6                 & 42.8               & 60.6                       & 38.7         & 54.7            & 79.9              \\
             & 75                        & 39.2                 & 50.4               & 69.0                         & 46.4         & 62.1            & 79.9              \\
             & 90                         & 36.2                 & 50.2               & 58.2                       & 44.5         & 60.8            & 85.4               \\
             & 100                           & 13.6                 & 26.0                 & 58.2                       & 18.2         & 26.6            & 52.8              \\ \hline
\multirow{6}{*}{FairytaleQA}      & 0                           & 0.0                    & 0.0                  & 0.0                          & 39.3         & 55.3             & 90.8              \\
      & 25                        & 0.0                    & 0.0                  & 0.0                          & 40.0         & 56.1            & 90.5               \\
      & 50                         & 0.0                    & 0.0                  & 0.0                          & 39.2         & 55.3             & 88.9               \\
      & 75                        & 0.0                    & 0.0                  & 0.0                          & 39.7         & 55.4            & 90.6              \\
      & 90                         & 0.0                    & 0.0                  & 0.0                          & 38.1          & 54.2             & 89.9               \\
      & 100                           & 0.0                    & 0.0                  & 0.0                          & 30.1         & 49.5            & 88.4              \\ \hline
\multirow{6}{*}{QAConv}           & 0                           & 29.4                & 36.3               & 49.4                      & 35.1         & 51.6            & 89.9              \\
           & 25                        & 28.6                & 34.5               & 48.0                      & 33.5         & 48.7            & 89.6              \\
           & 50                         & 28.0                & 34.0              & 47.1                       & 33.1          & 48.7            & 89.5              \\
           & 75                        & 28.6                & 35.9              & 48.5                      & 34.2         & 50.5            & 90.0              \\
           & 90                         & 29.0                & 36.0              & 49.9                      & 34.8         & 50.6            & 89.0              \\
           & 100                           & 23.5                 & 34.3              & 41.0                      & 30.2         & 45.2            & 87.2              \\ \hline
\multirow{6}{*}{ROPES}            & 0                           & 66.8                & 67.4              & 72.2                      & 67.0         & 72.7            & 96.0              \\
            & 25                        & 66.8                & 67.5              & 69.8                      & 67.2         & 71.1            & 96.2              \\
            & 50                         & 62.8                & 63.4              & 65.5                      & 63.1         & 66.6            & 95.2              \\
            & 75                        & 66.8                & 68.0                 & 68.8                     & 67.4         & 70.8            & 96.2               \\
            & 90                         & 70.6                & 71.5              & 71.8                      & 71.0         & 73.0            & 96.8              \\
            & 100                           & 60.8                & 63.9              & 61.2                      & 62.1         & 64.9             & 95.3              \\ \bottomrule
\end{tabular}
\caption{Full Results for the QA datasets. There is overwhelming agreement between all metrics on the ranking between models trained on different synthetic fractions. EM: Exact Match, Inc: String Inclusion, R Inc: Reverse String Inclusion}
\label{table:qafull}
\end{table*}

\begin{table*}[]
\centering
\begin{tabular}{@{}lllrrr@{}}
\toprule 
\textbf{Run} & \textbf{Dataset}     & \textbf{Synthetic \%} & \textbf{BLEU}  & \textbf{ROUGE} & \textbf{BERTScore}  \\ \midrule
0   & \multirow{3}{*}{FairytaleQA} & 95           & 38.78 & 54.80  & 90.34 \\
   &  & 97.5          & 37.19 & 52.09 & 86.17 \\
   &  & 100              & 26.10  & 43.82 & 77.03 \\ \midrule
   & \multirow{3}{*}{QAConv}      & 95           & 34.45 & 51.23 & 90.35 \\
   &       & 97.5          & 34.31 & 51.89 & 89.80  \\
   &       & 100              & 32.33 & 49.39 & 89.77 \\ \midrule
   & \multirow{3}{*}{CoQA}        & 95           & 25.33 & 35.64 & 59.39 \\
   &         & 97.5          & 42.78 & 57.88 & 78.84 \\
   &         & 100              & 19.11 & 27.57 & 51.44 \\ \midrule
   & \multirow{3}{*}{ROPES}       & 95           & 72.89 & 77.26 & 97.22 \\
   &        & 97.5          & 70.74 & 73.83 & 96.58 \\
   &        & 100              & 58.28 & 60.82 & 94.37 \\ \midrule \midrule
1   & \multirow{3}{*}{FairytaleQA} & 95           & 39.50  & 54.67 & 90.57 \\
   &  & 97.5          & 35.95 & 53.10  & 89.73 \\
   &  & 100              & 29.75 & 47.05 & 81.86 \\ \midrule
   & \multirow{3}{*}{QAConv}      & 95           & 38.94 & 56.41 & 90.79 \\
   &       & 97.5          & 40.06 & 57.64 & 90.45 \\
   &       & 100              & 37.46 & 54.11 & 88.57 \\ \midrule
   & \multirow{3}{*}{CoQA}        & 95           & 34.84 & 47.16 & 63.35 \\
   &         & 97.5          & 41.91 & 56.53 & 85.30  \\
   &         & 100              & 14.81 & 24.58 & 50.02 \\ \midrule
   & \multirow{3}{*}{ROPES}       & 95           & 69.61 & 71.97 & 96.23 \\
   &        & 97.5          & 69.72 & 72.80  & 96.58 \\
   &        & 100              & 62.08 & 65.44 & 95.03 \\ \midrule \midrule
2   & \multirow{3}{*}{FairytaleQA} & 95           & 37.97 & 53.98 & 90.34 \\
   &  & 97.5          & 37.65 & 52.44 & 87.83 \\
   &  & 100              & 29.26 & 49.70  & 88.56 \\ \midrule
   & \multirow{3}{*}{QAConv}      & 95           & 38.07 & 54.49 & 90.13 \\
   &       & 97.5          & 37.70  & 54.64 & 88.61 \\
   &       & 100              & 35.94 & 51.84 & 89.49 \\ \midrule
   & \multirow{3}{*}{CoQA }       & 95           & 30.80  & 42.03 & 62.72 \\
   &         & 97.5          & 30.40  & 41.02 & 56.83 \\
   &         & 100              & 23.42 & 37.20  & 63.31 \\ \midrule
   & \multirow{3}{*}{ROPES}       & 95           & 63.46 & 65.80  & 95.28 \\
   &        & 97.5          & 67.94 & 71.47 & 96.16 \\
   &        & 100              & 58.34 & 61.62 & 94.16 \\ \bottomrule
\end{tabular}
\caption{Results on $n=5000$ from 95\% to 100\% for the QA datasets. There is overwhelming agreement between all metrics on the ranking between models trained on different synthetic fractions.}
\label{table:qaz5}
\end{table*}

\begin{table*}[ht]
\centering
\begin{tabular}{@{}lllr@{}}
\toprule
Train Sets     & Test Set & Synthetic \% & Test Accuracy \\ \midrule
\multirow{5}{*}{FEVER, SciFact} & \multirow{5}{*}{WANLI}    & 0                    & 69.98         \\
               &          & 25                   & 67.56         \\
               &          & 50                   & 65.86         \\
               &          & 75                   & 64.82         \\
               &          & 100                  & 64.36         \\\midrule
\multirow{5}{*}{WANLI, SciFact} & \multirow{5}{*}{FEVER}    & 0                    & 83.01         \\
               &          & 25                   & 80.64         \\
               &          & 50                   & 79.22         \\
               &          & 75                   & 78.94         \\
               &          & 100                  & 76.01         \\\midrule
\multirow{5}{*}{FEVER, WANLI}   & \multirow{5}{*}{SciFact}  & 0                    & 71.76         \\
               &          & 25                   & 69.75         \\
               &          & 50                   & 69.82         \\
               &          & 75                   & 66.42         \\
               &          & 100                  & 64.41         \\ \bottomrule
\end{tabular}
\caption{Test accuracy when replacing human data with synthetic data in the out-of-distribution setting. Using more synthetic data leads to performance declines even in the OOD setting, showing that human data is of higher quality and the results from the main text cannot be explained by a spurious correlation between the human test and human training samples.}
\label{table:distshift}
\end{table*}

\begin{table*}[ht]
\centering
\begin{tabular}{@{}lrr@{}}
\toprule
\multirow{2}{*}{Dataset} & \multicolumn{2}{r}{Claim / Question} \\
         & Synthetic  & Human  \\ \midrule
FEVER       & 35.78                      & 42.76                  \\
WANLI       & 15.15                      & 20.10                   \\
SCIFACT     & 7.12                       & 20.92                  \\
FACTIFY     & 14.50                       & 23.93                  \\ \midrule
NarrativeQA & 30.92                      & 8.25                   \\
CoQA        & 7.08                       & 8.39                   \\
FairyTaleQA & 22.59                      & 16.85                  \\
ROPES       & 28.73                      & 41.42                  \\ \bottomrule
\end{tabular}
\caption{4-Gram overlap \% between all synthetic and human generated claims / questions for each dataset. On several datasets, synthetic claims have a lower overlap}
\label{table:ngramoverlap}
\end{table*}

\end{document}